\documentclass{article}

\usepackage{microtype}
\usepackage{graphicx}
\usepackage{booktabs} % for professional tables

\usepackage{hyperref}

\usepackage[accepted]{icml2025_arxiv}
\usepackage{amsmath}
\usepackage{amssymb}
\usepackage{mathtools}
\usepackage{amsthm}

\usepackage[capitalize,noabbrev]{cleveref}
\usepackage{bm,booktabs,float,subcaption}

\theoremstyle{plain}

\theoremstyle{definition}

\theoremstyle{remark}

\usepackage[textsize=tiny]{todonotes}

\icmltitlerunning{Adaptive Data Exploitation in Deep Reinforcement Learning}

\begin{document}

\twocolumn[
\icmltitle{
% ADEPT: Automatic Data Efficiency Tuning in Deep Reinforcement Learning
% AUDIT: Boosting Data Efficiency and Generalization in Deep Reinforcement Learning
Adaptive Data Exploitation in Deep Reinforcement Learning
}
% AUDIT: A Practical Solution for Data-Efficient and Generalizable Reinforcement Learning

% AUDIT: \textit{Au}tomatic \textbf{D}ata Ut\textbf{I}lization \textbf{T}uning



\icmlsetsymbol{equal}{*}

\begin{icmlauthorlist}
\icmlauthor{Mingqi Yuan}{comp}
\icmlauthor{Bo Li}{comp}
\icmlauthor{Xin Jin}{eit}
\icmlauthor{Wenjun Zeng}{eit}
\end{icmlauthorlist}

\icmlaffiliation{comp}{Department of Computing, The Hong Kong Polytechnic University, Hong Kong, China}
\icmlaffiliation{eit}{Ningbo Institute of Digital Twin, Eastern Institute of Technology, Ningbo, Zhejiang, China}

\icmlcorrespondingauthor{Xin Jin}{jinxin@eitech.edu.cn}

\icmlkeywords{Machine Learning, ICML}

\vskip 0.3in
]



\printAffiliationsAndNotice{\icmlEqualContribution} % otherwise use the standard text.

\begin{abstract}
We introduce \textbf{ADEPT}: \textbf{A}daptive \textbf{D}ata \textbf{E}x\textbf{P}loi\textbf{T}ation, a simple yet powerful framework to enhance the \textbf{data efficiency} and \textbf{generalization} in deep reinforcement learning (RL). Specifically, ADEPT adaptively manages the use of sampled data across different learning stages via multi-armed bandit (MAB) algorithms, optimizing data utilization while mitigating overfitting. Moreover, ADEPT can significantly reduce the computational overhead and accelerate a wide range of RL algorithms. We test ADEPT on benchmarks including Procgen, MiniGrid, and PyBullet. Extensive simulation demonstrates that ADEPT can achieve superior performance with remarkable computational efficiency, offering a practical solution to data-efficient RL. Our code is available at \url{https://github.com/yuanmingqi/ADEPT}.

% We present \textbf{ADEPT}: \textbf{A}utomatic \textbf{D}ata \textbf{E}fficiency \textbf{T}uning, a simple and effective framework to enhance the data efficiency and generalization in deep reinforcement learning (RL). Specifically, ADEPT adaptively controls the extent of utilization of the sampled data at different learning stages, which optimizes the data utilization and prevents the agent from overfitting. Moreover, ADEPT can significantly reduce the computational overhead and accelerate a wide range of RL algorithms. We test ADEPT on various tasks of Procgen, MiniGrid, and Pybullet. Extensive simulation demonstrates that ADEPT can achieve superior performance with remarkable computational efficiency, providing a practical solution to data-efficient RL. %Our code is available at \url{github.com}.

% for accelerating reinforcement learning (RL) algorithms. While RL has achieved remarkable success, its high computational cost and data inefficiency remain major challenges. ADEPT addresses these issues by dynamically tuning policy update epochs to balance exploration and exploitation, reducing computational overhead while enhancing performance. Experiments on the Procgen benchmark demonstrate that ADEPT significantly improves training efficiency, sample utilization, and generalization compared to fixed update schedules, offering a practical solution for data-efficient RL.
\end{abstract}

\section{Introduction}
Deep reinforcement learning (RL) has achieved remarkable success in diverse domains such as complex games \cite{mnih2015human,silver2016mastering,vinyals2019grandmaster}, algorithm innovation \cite{fawzi2022discovering,mankowitz2023faster}, large language model (LLM) \cite{ouyang2022training}, and chip design \cite{goldie2024addendum}. However, promoting \textbf{data efficiency} and \textbf{generalization} remains a long-standing challenge in deep RL, especially when learning from complex environments with high-dimensional observations ({\em e.g.}, images). The agents often require millions of interactions with their environment, resulting in substantial computational overhead and restricting applicability in scenarios where data collection is costly or impractical. Moreover, the learned policy frequently struggles to adapt to dynamic environments in which minor variations in tasks or conditions may significantly degrade its performance.

\begin{figure}[t]
% \vskip 0.1in
\begin{center}
\centerline{\includegraphics[width=\linewidth]{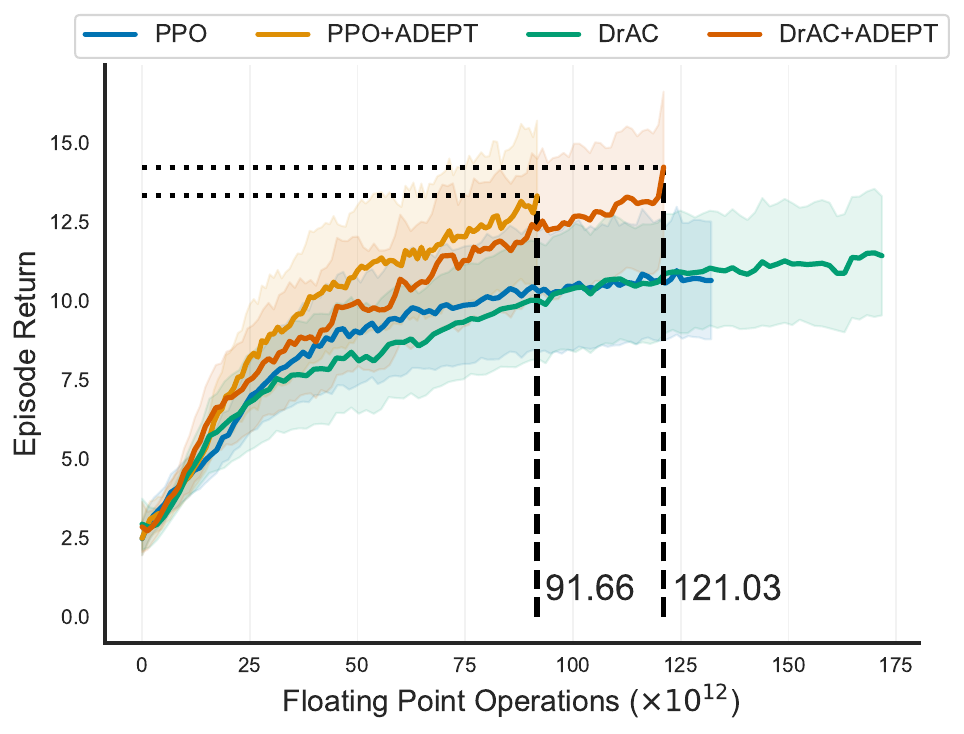}}
\caption{Aggregated training performance and computational overhead comparison on the Procgen benchmark. ADEPT serves as a plug-and-play module to enhance RL algorithms, which can significantly promote data efficiency and reduce computational costs.}
\label{fig:preface}
\end{center}
\vskip -0.4in
\end{figure}

To tackle the data efficiency problem, research has primarily focused on two key aspects: \textbf{data acquisition} and \textbf{data exploitation}. Data acquisition aims to maximize the quality and diversity of the data collected during interactions with the environment, while data exploitation seeks to optimize the utility of this data to enhance learning efficiency. Prominent techniques include data augmentation \cite{pathak2017curiosity,yuan2022rewarding,henaff2022exploration,yuan2023automatic,laskin2020reinforcement,laskin2020curl}, efficient experience replay \cite{schaul2016prioritized,horgan2018distributed,andrychowicz2017hindsight}, distributed training \cite{espeholt2018impala,barth2018distributed}, and environment acceleration \cite{petrenko2020sample,makoviychuk2isaac,weng2022envpool}. For example, \cite{burda2018exploration} proposed RND that utilizes the prediction error against a fixed network as intrinsic rewards, enabling structured and efficient exploration. In contrast, RAD \cite{laskin2020reinforcement} applies simple transformations to input observations during training, improving both data efficiency and generalization. On the exploitation side, \cite{schaul2016prioritized} developed prioritized experience replay (PER), which enhances learning efficiency by sampling high-priority experiences more frequently during training. However, these methods struggle to balance data efficiency and computational efficiency. Techniques like data augmentation often introduce auxiliary models, storage, and optimization objectives, which significantly increase the computational overhead \cite{badia2020never,yarats2021image}. Moreover, they often lack guarantees for optimizing long-term returns. For instance, intrinsic reward approaches suffer from the policy-invariant problem, where the exploration incentivized by intrinsic rewards may diverge from the optimal policy. 

Beyond data efficiency, generalization represents another critical challenge in deep RL, as agents often overfit to their training environments \cite{kuttler2020nethack,andrychowicz2021what,raileanu2021decoupling}. To address this issue, \cite{cobbe2021phasic} proposed a phasic policy gradient (PPG), which explores decoupling the representation learning of policy and value networks. PPG also incorporates an auxiliary learning phase to distill the value function and constrain the policy, thereby enhancing generalization. Similarly, \cite{raileanu2021decoupling} proposed decoupled advantage actor-critic (DAAC), which also leverages decoupled policy and value networks but eliminates the need for auxiliary learning. Both PPG and DAAC can achieve superior generalization performance on the Procgen benchmark \cite{cobbe2020leveraging}. However, they rely on sophisticated hyperparameter tuning ({\em e.g.}, the number of update epochs of the two separated networks), which limits their adaptability across diverse scenarios. Moreover, they lack mechanisms to dynamically adjust to different learning stages, further constraining their applicability.

% Finally, regarding the environment aspect, \cite{weng2022envpool} developed EnvPool, a high-performance environment simulator that significantly accelerates data collection by parallelizing environment interactions at scale. EnvPool can drastically reduce the time required to train RL agents and facilitate a wide range of RL libraries.
% Despite the achievements of the methods above, there are still several limitations. First, these methods are unable to strike a balance between data efficiency and computational efficiency. Techniques like data augmentation often introduce auxiliary models, storage, operations, and optimization objectives, which significantly increase the computational overhead \cite{badia2020never,yarats2021image}. Moreover, they lack systematic guarantees for maximizing long-term cumulative returns. For instance, PER focuses on samples with high temporary difference errors to accelerate learning but may overemphasize short-term corrections, potentially neglecting transitions that are critical for global policy improvement \cite{schaul2016prioritized}. Similarly, intrinsic reward approaches suffer from the policy-invariant problem, where the exploration incentivized by intrinsic rewards may fail to align with the true optimal policy. As for the environmental side, it cannot cannot fundamentally affect the algorithms.

Inspired by the discussions above, we aim to enhance the data efficiency and generalization of RL agents while minimizing computational overhead. To that end, we propose a novel framework entitled \textbf{ADEPT}: \textbf{A}daptive \textbf{D}ata \textbf{E}x\textbf{P}loi\textbf{T}ation, which incorporates three scheduling algorithms to assist RL algorithms. Our main contributions are summarized as follows:
\begin{itemize}
    \item We propose to adaptively control the utilization of the sampled data across different tasks and learning stages. This scheduling process is formulated as a multi-armed bandit (MAB) problem, in which a set of extent values represents the arms. ADEPT can automatically select the optimal extent value based on the estimated task return, significantly maximizing the data efficiency while reducing the computational costs;
    
    \item By adaptively adjusting data utilization, ADEPT can effectively prevent the RL agent from overfitting and enhance its generalization ability. In particular, ADEPT has a simple architecture and requires no additional learning processes, which can facilitate a wide range of RL algorithms;
    
    \item Finally, we evaluate ADEPT on Procgen (sixteen procedurally-generated environments), MiniGrid (environments with sparse rewards), and PyBullet (robotics environments with continuous action space). Extensive simulation results demonstrate that ADEPT can achieve superior performance with remarkable computational efficiency. 
\end{itemize}
\section{Related Work}
\subsection{Data Efficiency in RL}
Observation augmentation and intrinsic rewards have emerged as promising approaches to promoting data efficiency in RL. \cite{yarats2021image} proposed data-regularized Q (DrQ) that leverages standard image transformations to perturb input observations and regularize the learned value function. DrQ enables robust learning directly from pixels without using auxiliary losses or pre-training, which can be combined with various model-free RL algorithms. \cite{yarats2021mastering} further extended DrQ and proposed DrQ-v2, which is the first model-free RL algorithm that solves complex humanoid locomotion tasks directly from pixel observations. In contrast, \cite{yuan2022renyi} proposed RISE that maximizes the R\'enyi entropy of the state visitation distribution and transforms the estimated sample mean into particle-based intrinsic rewards. RISE can achieve significant exploration diversity without using any auxiliary models.

In this paper, we improve the data efficiency from the perspective of exploitation. Our method provides a systematic guarantee for optimizing long-term returns and significantly reduces computational costs.

\begin{figure*}[t!]
% \vskip 0.2in
\begin{center}
\centerline{\includegraphics[width=\linewidth]{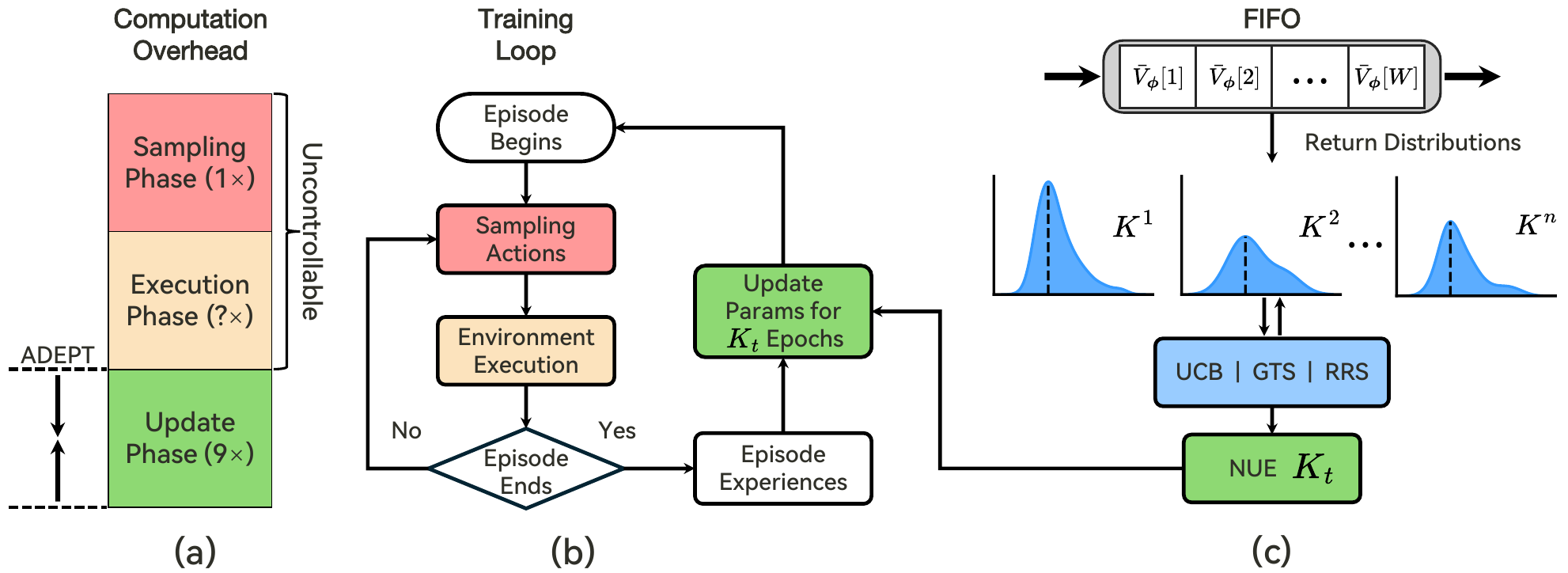}}
\caption{Overview of the ADEPT framework. (a) The proportion of the computational overhead (FLOPS) is evaluated using CleanRL's PPO implementation \cite{huang2022cleanrl} and the Procgen benchmark \cite{cobbe2020leveraging}. Since the overhead of the execution
phase depends on various factors, it is not counted here. (b) A typical workflow of the on-policy RL algorithms. (c) ADEPT optimizes data utilization by adjusting the number of update epochs (NUE) in the update phase.}
\label{fig:overview}
\end{center}
\vskip -0.35in
\end{figure*}

\subsection{Generalization in RL}
Achieving robust generalization is a fundamental challenge in RL. To that end, extensive research has focused on techniques such as regularization \cite{srivastava2014dropout,bengio2017deep}, data augmentation \cite{raileanu2020automatic,laskin2020reinforcement,ye2020rotation,wang2020improving}, representation learning \cite{igl2019generalization,laskin2020curl,sonar2021invariant,stooke2021decoupling}, representation decoupling \cite{raileanu2021decoupling,cobbe2021phasic}, causal modeling \cite{mutti2023provably}, exploration \cite{jiang2024importance}, and gradient strategies \cite{liu2023improving}. For instance, \cite{moon2022rethinking} proposed delayed-critic policy gradient (DCPG), which mitigates overfitting and enhances observational generalization by optimizing the value network less frequently but with larger datasets. In contrast, \cite{liu2023improving} proposes a conflict-aware gradient agreement augmentation (CG2A) framework that combines data augmentation techniques with gradient harmonization. CG2A improves generalization and sample efficiency by resolving gradient conflicts and managing high-variance gradients.

In this paper, we improve the generalization by adaptively managing the utilization of sampled data without introducing any auxiliary learning.

\subsection{MAB Algorithms for RL}
MAB problems are closely related to RL, as both address decision-making under uncertainty \cite{auer2002finite}. While RL focuses on sequential decisions to maximize cumulative rewards, MAB methods optimize immediate actions, making them well-suited for tackling subproblems within RL frameworks, such as exploration ({\em e.g.}, $\epsilon$-greedy and Boltzmann exploration) \cite{sutton2018reinforcement} and dynamic resource allocation \cite{whittle1988restless}. For example, \cite{raileanu2020automatic} proposed UCB-DrAC, which employs a bandit algorithm to select optimal data augmentations, significantly improving generalization in RL. Similarly, \cite{badia2020agent57} designed Agent57 that uses a meta-controller based on bandit principles to balance exploration and exploitation across a family of policies, achieving human-level performance on all Atari games. Finally, AIRS \cite{yuan2023automatic} formulates intrinsic reward selection as a bandit problem, dynamically adapting rewards to enhance exploration at different learning stages.

In this paper, we leverage MAB algorithms to schedule the update phase of RL algorithms, optimizing data utilization and mitigating overfitting.

\section{Background}
\subsection{Reinforcement Learning}
We frame the RL problem considering a Markov decision process (MDP) \citep{bellman1957markovian, kaelbling1998planning} defined by a tuple $\mathcal{M}=(\mathcal{S},\mathcal{A},r,P,d_{0},\gamma)$, where $\mathcal{S}$ is the state space, $\mathcal{A}$ is the action space, and $r:\mathcal{S}\times\mathcal{A}\rightarrow\mathbb{R}$ is the extrinsic reward function, $P:\mathcal{S}\times\mathcal{A}\rightarrow\Delta(\mathcal{S})$ is the transition function that defines a probability distribution over $\mathcal{S}$, $d_{0}\in\Delta(\mathcal{S})$ is the distribution of the initial observation $\bm{s}_{0}$, and $\gamma\in[0, 1)$ is a discount factor. The goal of RL is to learn a policy $\pi_{\bm\theta}(\bm{a}|\bm{s})$ to maximize the expected discounted return:
	\begin{equation}
		J_{\pi}(\bm{\theta})=\mathbb{E}_{\pi}\left[\sum_{t=0}^{\infty}\gamma^{t}r_{t}\right].
	\end{equation}

\subsection{Workflow of On-policy RL Algorithms}
Figure~\ref{fig:overview}(b) illustrates a typical workflow of on-policy RL algorithms. In each episode, the agent samples actions based on its current policy, which are then executed by the environment. At the end of the episode, the agent leverages the accumulated experiences to update its parameters for a certain number of epochs. The action sampling, environment execution, and model update take up most of the computational costs, whose broad proportion is illustrated in Figure~\ref{fig:overview}(a).

\section{Adaptive Data Exploitation}
In this section, we propose the ADEPT framework to enhance data efficiency and generalization while reducing computational overhead during training. Our key insights are threefold: (i) Prior work \cite{cobbe2021phasic,raileanu2021decoupling} indicates that different tasks benefit from varying levels of data utilization, which is directly controlled by the number of update epochs (NUE). As shown in Figure~\ref{fig:overview}, NUE values reflect how much the learning process relies on sampled data. By adaptively adjusting NUE, we can better align with the dynamic nature of learning, particularly in procedurally-generated environments. (ii) A dynamic NUE value can reduce reliance on specific data and preserve the agent's plasticity throughout the training process, preventing overfitting and thereby improving generalization. (iii) Among the three key phases in Figure~\ref{fig:overview}(a), the model update phase incurs the highest computational overhead. By minimizing unnecessary updates through adaptive NUE tuning, we can significantly reduce the overall computational overhead. This approach allows us to concentrate computational resources on the most impactful updates, leading to a more efficient training process.

% Among the three key phases in Figure~\ref{fig:overview}(a), only the computational overhead of the model update phase is directly influenced by the RL algorithm. The other two phases depend heavily on the environment and vary significantly. Therefore, optimizing the model update phase by adaptively tuning NUE values is an effective way to reduce overall computational costs. 

Denote by $\mathcal{K}=\{K^{1}, K^{2},\dots, K^{n}\}$ the set of NUE values, the value selection at different learning stages can be formulated as a MAB problem. Each value is considered an arm, and the objective is to maximize the long-term return evaluated by the task reward function. In the following, we introduce three specific algorithms to solve the defined MAB problem.

\subsection{Upper Confidence Bound}
We first leverage the upper confidence bound (UCB) \cite{auer2002using} to solve the defined MAB problem, which is a representative and effective method. Specifically, UCB selects actions by the following policy:
\begin{equation}\label{eq:ucb}
    K_{t}=\underset{K\in\mathcal{K}}{\rm argmax}\left[Q_{t}(K)+c\sqrt{\frac{\log t}{N_{t}(K)}}\right],
\end{equation}
where $K_{t}$ is the NUE value selected at time step $t$, $N_{t}(K)$ is the number of times that $K$ has been chose before time step $t$, and $c$ is the exploration coefficient. Before the $t$-th update, we select a $K$ using Eq.~(\ref{eq:ucb}), which will used for the policy updates. Then, the counter is updated by $N_t(K)=N_t(K)+1$. Next, we collect rollouts with the new policy and update the Q-function using a sliding window average of the past mean returns obtained by the agent after being updated using $K$:
\begin{equation}
    Q_{t}(K)=\frac{1}{W}\sum_{i=1}^{W}\bar{V}_{\bm \phi}[i],
\end{equation}
where $\bar{V}_{\bm \phi}$ is the average estimated task return of the episode.

UCB encourages the exploration of less-used $K$ values while progressively focusing on the most promising options. We refer to this algorithm as \textbf{ADEPT(U)}.

\subsection{Gaussian Thompson Sampling}
Furthermore, we introduce Gaussian Thompson sampling (GTS) \cite{thompson1933likelihood} to solve the same MAB problem by modeling the return distribution of each $K$ as a Gaussian distribution. At each time step, GTS samples from the distributions corresponding to all NUE candidates and selects the one with the highest sampled value:

\begin{equation}\label{eq:gts}
    K_{t} = \underset{K \in \mathcal{K}}{\rm argmax} \ \mathcal{N}\left(\mu_t(K), \sigma_t^2(K)\right),
\end{equation}

where \( \mu_t(K) \) and \( \sigma_t^2(K) \) are the mean and variance of the reward distribution for \( K \). Then the parameters are updated incrementally as follows:
\begin{equation}
\begin{aligned}
        \mu_{t+1}(K) &= \mu_t(K) + \eta \cdot  \frac{Q_t(K) - \mu_t(K)}{N_t(K) + 1}, \\
    \sigma_{t+1}^2(K) &= \frac{N_t(K) \sigma_t^2(K) + (Q_t(K) - \mu_t(K))^2}{N_t(K) + 1},
\end{aligned}
\end{equation}
where $\eta$ is a step size. 

GTS allows for a more flexible exploration pattern that adapts dynamically to new information compared to the fixed confidence bound strategy in UCB. We refer to this algorithm as \textbf{ADEPT(G)}.

\begin{algorithm}[t!]
	\caption{Adaptive Data Exploitation (UCB)}
	\label{algo:autodet(u)}
	\begin{algorithmic}[1]
    \STATE Initialize the policy network $\pi_{\bm \theta}$ and value network $V_{\bm \phi}$;
    \STATE Initialize a set $\mathcal{K}$ of NUE values, an exploration coefficient $c$, a window length $W$ for estimating the Q-functions;
    \STATE $\forall K\in\mathcal{K}$, let $$N(K)=1, Q(K)=0, R(K)={\rm FIFO}(W){\rm ;}$$
    \FOR{each episode $e$}{
        \STATE Sample rollouts using the policy network $\pi_{\bm \theta}$;
        \STATE Perform the generalized advantage estimation (GAE) to get the estimated returns;
        \STATE Select $K_e$ using Eq.~(\ref{eq:ucb});
        \STATE Update policy network and value network;
        \STATE Compute the mean return $\bar{V}_{\bm\phi}$ obtained by the new policy;
        \STATE Add $\bar{V}_{\bm\phi}$ to the queue $R(K_{e})$ using the first-in-first-out rule;
        \STATE $Q(K_e)\leftarrow\frac{1}{|R(K_e)|}\sum_{\bar{V}_{\bm\phi}\in R(K_{e})}\bar{V}_{\bm\phi}$;
        \STATE $N(K_{e})\leftarrow N(K_{e})+1$;
        }
	\ENDFOR
	\end{algorithmic}
\end{algorithm}

% \begin{figure*}[b!]
% \vskip 0.2in
% \begin{center}
% \centerline{\includegraphics[width=\linewidth]{figures/pg_ppo_st8.pdf}}
% \caption{Performance and time cost comparison of the vanilla PPO and PPO+ADEPT on eight Procgen environments. The solid line and shaded regions represent the mean and standard deviation, respectively, across five runs. Note that the dotted line and dashed line represent the highest score and the shortest time cost, respectively.}
% \label{fig:pg_ppo_scores_and_times_8}
% \end{center}
% \vskip -0.2in
% \end{figure*}

% \begin{figure*}[t!]
% \vskip 0.2in
% \begin{center}
% \centerline{\includegraphics[width=\linewidth]{figures/pg_ppo_st8.pdf}}
% \caption{Performance .}
% \label{fig:pg_daac_scores_and_times_8}
% \end{center}
% \vskip -0.2in
% \end{figure*}
 
\subsection{Round-Robin Scheduling}
Finally, we employ Round-Robin scheduling (RRS) \cite{arpaci2018operating} to ensure that each candidate in \( \mathcal{K} \) is selected in a cyclical order, giving equal opportunity to all options without bias. This strategy is widely used in various domains, such as network packet scheduling, load balancing in distributed systems, and time-sharing in resource management. 
% Similarly, in our context, it ensures that all NUE candidates are explored fairly and periodically, avoiding the risk of neglecting any option.

The selection at time step \( t \) follows:

\begin{equation}
    K_{t} = \mathcal{K}\left[ (t \bmod |\mathcal{K}|) + 1 \right],
\end{equation}

where \( |\mathcal{K}| \) is the cardinality of the set $\mathcal{K}$. This algorithm is referred to as \textbf{ADEPT(R)}.

\section{Experiments}
\begin{figure*}[t!]
% \vskip 0.1in
\centering
\subcaptionbox{\textit{PPO v.s. PPO+ADEPT}}{\includegraphics[width=\linewidth]{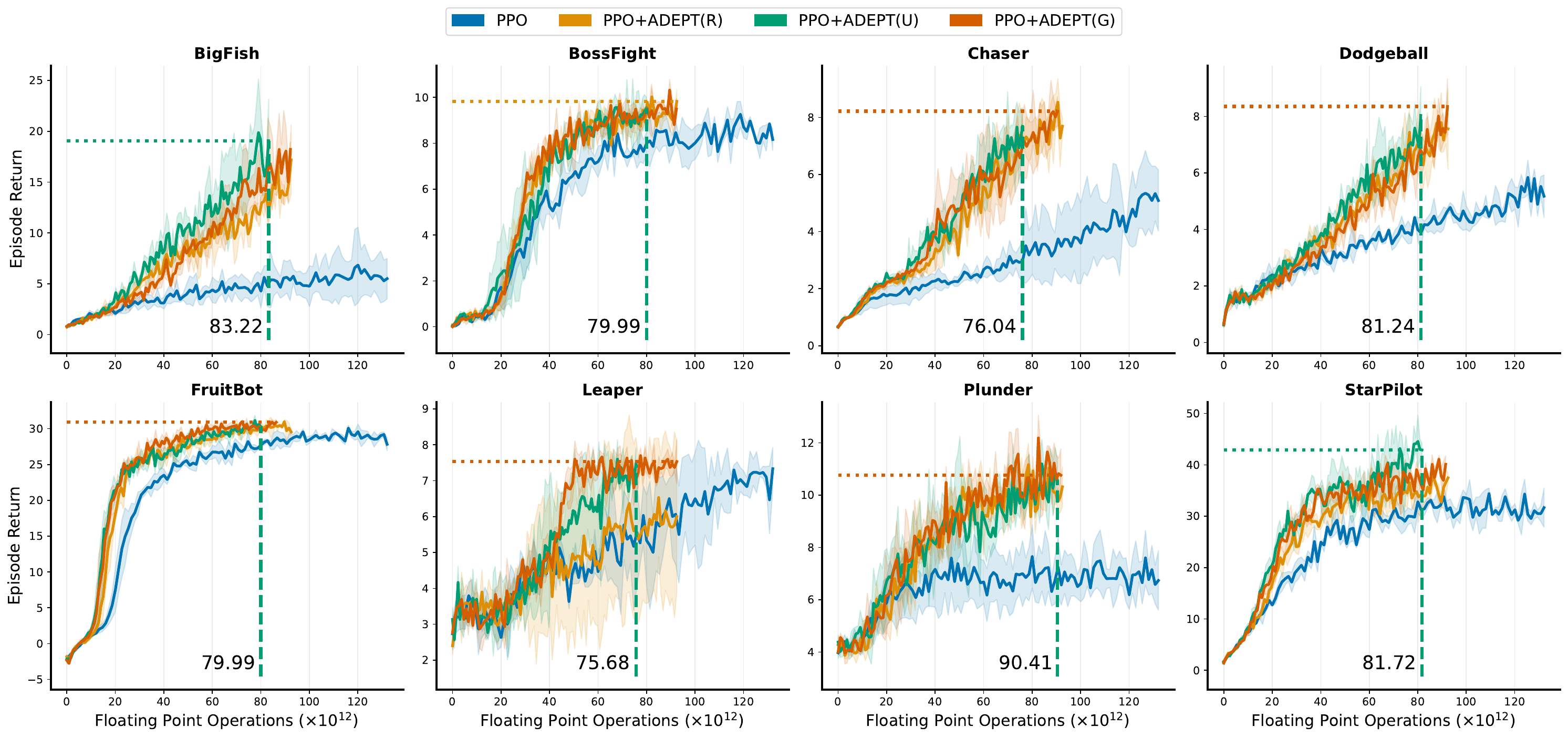}}
\subcaptionbox{\textit{DrAC v.s. DrAC+ADEPT}}{\includegraphics[width=\linewidth]{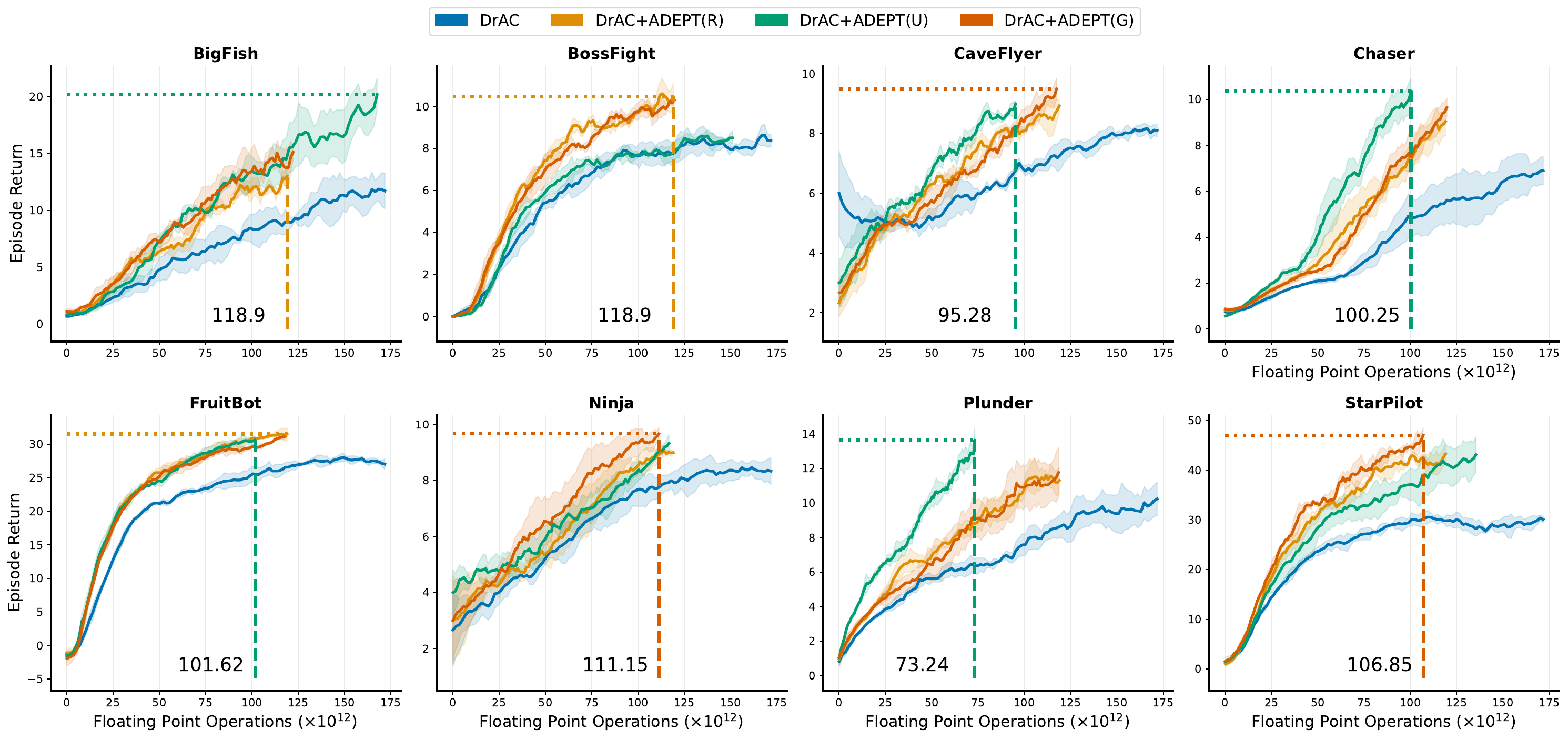}}
\caption{Training performance and computational overhead comparison of the PPO, DrAC, and their combinations with ADEPT on eight Procgen environments. The solid line and shaded regions represent the mean and standard deviation, respectively, across five runs. Note that the dotted line and dashed line represent the highest score and the lowest overhead, respectively.}
\label{fig:pg_ppo_drac_st_8}
\vskip -0.2in
\end{figure*}

In this section, we design the experiments to investigate the following questions:
\begin{itemize}
    \item \textbf{Q1}: Can ADEPT improve data efficiency as compared to using fixed NUE values? (See Figure~\ref{fig:preface}, \ref{fig:pg_ppo_drac_st_8}, \ref{fig:pb_ppo}, \ref{fig:mg_ppo_agg}, \ref{fig:pg_ppo_st16}, and \ref{fig:pg_drac_st16})
    \item \textbf{Q2}: Can ADEPT reduce the overall computational overhead? (See Figure~\ref{fig:preface}, \ref{fig:pg_ppo_drac_st_8}, \ref{fig:pg_ppo_st16}, and \ref{fig:pg_drac_st16})
    \item \textbf{Q3}: Can ADEPT achieve higher generalization performance in procedurally-generated environments? (See Figure~\ref{fig:pg_test_scores})
    \item \textbf{Q4}: What are the detailed decision processes of ADEPT? (See Figure~\ref{fig:pg_ppo_ucb_ts_decision_8}, \ref{fig:pg_ppo_ucb_decision_16}, \ref{fig:pg_ppo_ts_decision_16}, \ref{fig:pg_drac_ucb_decision_16}, and \ref{fig:pg_drac_ts_decision_16})
    \item \textbf{Q5}: How does ADEPT behave in sparse-rewards environments and continuous control tasks? (See Figure~\ref{fig:pb_ppo} and \ref{fig:mg_ppo_agg})
\end{itemize}

\subsection{Setup}
We first evaluate the ADEPT using the Procgen benchmark \cite{cobbe2020leveraging}, which contains sixteen procedurally-generated environments. We select Procgen for two reasons. First, Procgen is similar to the arcade learning environment (ALE) benchmark \cite{bellemare2013arcade} that requires the agent to learn motor control directly from images and presents a clear challenge to the agent's data efficiency. Moreover, Procgen provides procedurally generated levels to evaluate the agent's generalization ability with a well-designed protocol. All the environments use a discrete fifteen-dimensional action space and generate $(64,64,3)$ RGB observations. We use the \textit{easy} mode and train the agents on 200 levels before testing them on the full distribution of levels. Furthermore, we introduce the MiniGrid \cite{MinigridMiniworld23} and PyBullet \cite{coumans2016pybullet} to test ADEPT in sparse-rewards environments and continuous control tasks.

\textbf{Algorithmic Baselines}. For the algorithm baselines, we select the proximal policy optimization (PPO) \cite{schulman2017proximal} and data regularized actor-critic (DrAC) \cite{raileanu2020automatic} as the candidates. PPO is a representative algorithm that produces considerable performance on most existing RL benchmarks, while DrAC integrates data augmentation techniques into AC algorithms and significantly improves the agent's generalization ability in procedurally-generated environments. The details of the selected algorithmic baselines can be found in Appendix~\ref{appendix:baseline}.

\textbf{NUE Candidates}. \cite{cobbe2020leveraging} and \cite{raileanu2020automatic} reported the overall best hyperparameters for the two algorithms for the Procgen benchmark. Both PPO and DrAC update their parameters for 3 epochs after each episode, so we conduct experiments with $\mathcal{K}=\{3,2,1\}$ and $\mathcal{K}=\{5,3,2,1\}$. For the first set, we aim to assess whether ADEPT can enhance RL algorithms while reducing computational overhead. In contrast, the second set allows us to explore whether a broader range of NUE values further improves performance. For MiniGrid and PyBullet experiments, please refer to Appendix~\ref{appendix:exp setup}.

% PPO updates the parameters for 3 epochs after each episode, and the ADEPT(R) will select a value of $K_{t}$ from the set $\{3,2,1\}$ in turn. For DAAC, the general best number of update epochs for policy and value networks are 1 and 9, respectively. We follow the experiment setting of \cite{raileanu2021decoupling} and use $\{9,5,1\}, \{9,5\}$ when controlling the value network, and use $\{1,3,6\}, \{1,3\}$ when controlling the policy network. For MiniGrid and Pybullet experiments, please refer to Appendix~\ref{appendix:exp setup}.

\textbf{Evaluation Metrics}. We evaluate the data efficiency and generalization of each method using three metrics: (i) the average floating point operations (FLOPS) over 16 environments and all the runs, and the calculation process is depicted in Appendix~\ref{appendix:overhead}, (ii) the aggregated mean scores on 200 levels, and (iii) the aggregated mean, median, interquartile mean (IQM), and optimality gap (OG) \cite{agarwal2021deep} on the full distribution of levels. Note that the score of each method on each environment is computed as the average episode returns over 100 episodes and 5 random seeds.

More details about the experimental setup and hyperparameters selection can be found in Appendix~\ref{appendix:exp setup}.

\subsection{Results Analysis}
The following results analysis is performed based on the predefined research questions. We provide the detailed training curves of all the methods and configurations in Appendix~\ref{appendix:curves}.

\textbf{Data efficiency comparison}. Figure~\ref{fig:pg_ppo_drac_st_8} illustrates the data efficiency and computational overhead comparison between vanilla PPO, DrAC, and their combinations with three ADEPT algorithms in eight environments, with the full comparison provided in Appendix~\ref{appendix:efficiency}. By alternating NUE values from $\mathcal{K}$, PPO+ADEPT(R) achieves close or higher performance than the vanilla PPO agent, especially in the \textit{BigFish}, \textit{Chaser}, \textit{Dodgeball}, and \textit{Plunder} environments. The average computational overhead of PPO+ADEPT(R) is 70\% of the vanilla PPO agent. Similarly, PPO+ADEPT(U) outperforms the vanilla PPO agent in 14 environments and achieves the highest performance in 6 environments. Meanwhile, it produces the minimum computational overhead in 11 environments. In contrast, PPO+ADEPT(G) also achieves the highest performance in 6 environments and obtains the highest computational efficiency in 4 environments. Therefore, PPO+ADEPT(U) can achieve more extreme overhead compression than PPO+ADEPT(G).

\begin{figure*}[t!]
\vskip 0.1in
\begin{center}
\centerline{\includegraphics[width=\linewidth]{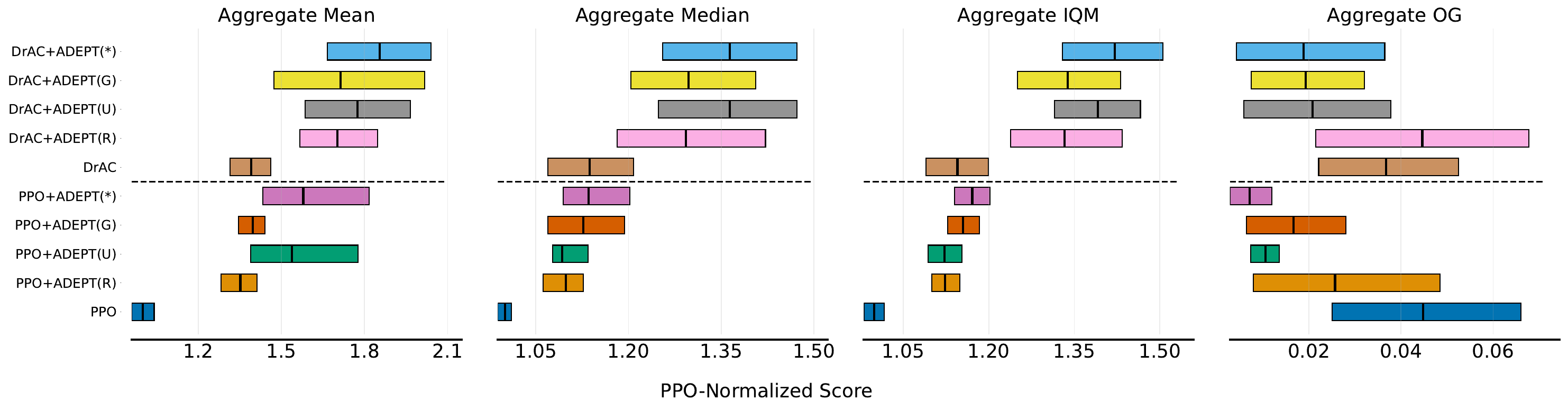}}
\caption{Aggregated performance of the PPO, DrAC, and their combinations with ADEPT on the test levels of the Procgen benchmark. All the scores are normalized by the corresponding PPO scores, and bars indicate $95\%$ confidence intervals computed using stratified bootstrapping over five random seeds. Note that $*$ represents the best scores gathered from all three ADEPT algorithms.}
\label{fig:pg_test_scores}
\end{center}
\vskip -0.2in
\end{figure*}

For the DrAC algorithm, DrAC+ADEPT(U) achieves the highest performance in 7 environments and obtains the highest computational efficiency in 6 environments. Specifically, DrAC+ADEPT(U) takes 69.1\% of the overhead to achieve the same or higher score in multiple environments against the vanilla DrAC agent, such as \textit{BossFight}, \textit{Chaser}, and \textit{CoinRun}. In contrast, DrAC+ADEPT(G) achieves the highest data efficiency in 5 environments and takes 68.7\% of the overhead of the vanilla DrAC agent. Finally, DrAC+ADEPT(R) also excels in 2 environments. The experiment results of PPO and DrAC demonstrate that ADEPT can significantly improve the data efficiency of RL algorithms and reduce the computational overhead.

\begin{figure}[h!]
\vskip 0.1in
\begin{center}
\includegraphics[width=\linewidth]{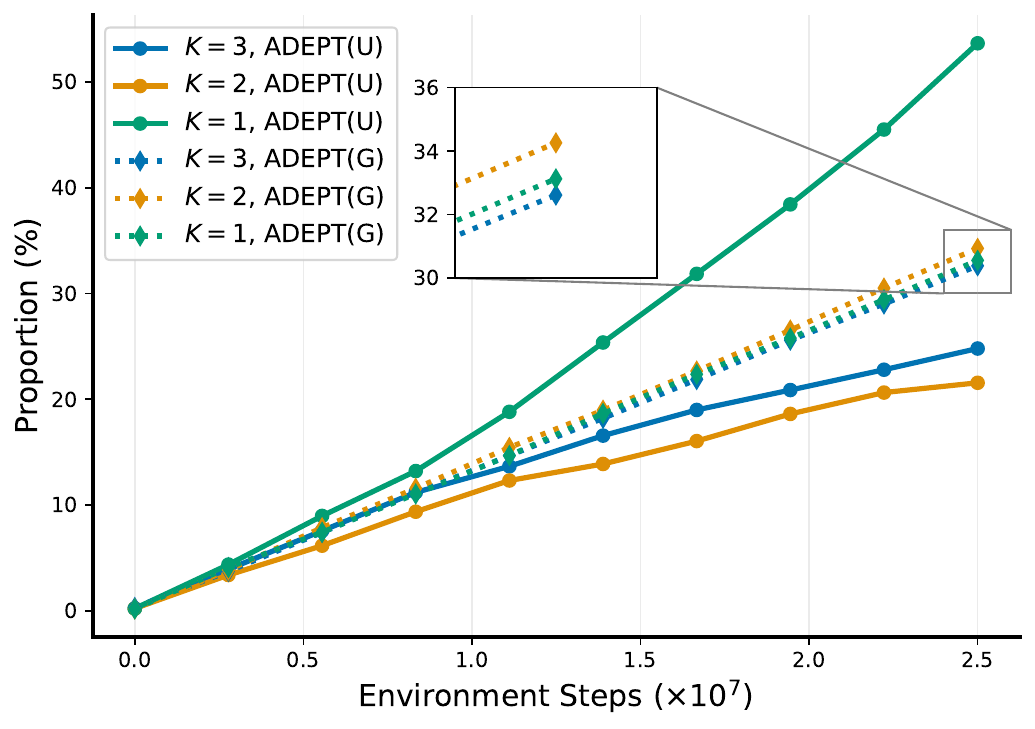}
\caption{The aggregated decision processes of ADEPT(U) and ADEPT(G) for PPO on the eight selected Procgen environments.}
\label{fig:pg_ppo_ucb_ts_decision_8}
\end{center}
\vskip -0.2in
\end{figure}

\textbf{Analysis of the decision process}. Next, we analyze the detailed decision processes of ADEPT. Figure~\ref{fig:pg_ppo_ucb_ts_decision_8} illustrates the cumulative proportion of each NUE value selected during the whole training of the PPO experiments. It is clear that ADEPT(U) primarily selects $K=1$, while $K=2$ and $K=3$ each account for approximately 20\%. In contrast, ADEPT(G) distributes its selections more evenly across all NUE values, reflecting a dynamic scheduling strategy enabled by its incremental update mechanism. These findings demonstrate that varying tasks and learning stages benefit from adaptive data utilization, and ADEPT can consistently select the most appropriate NUE values to maximize data efficiency. Additionally, ADEPT exhibits a similar pattern with DrAC as PPO. Finally, we provide the detailed decision processes of each method and environment in Appendix~\ref{appendix:decision}.

\begin{figure}[h!]
% \vskip 0.1in
\begin{center}
\includegraphics[width=\linewidth]{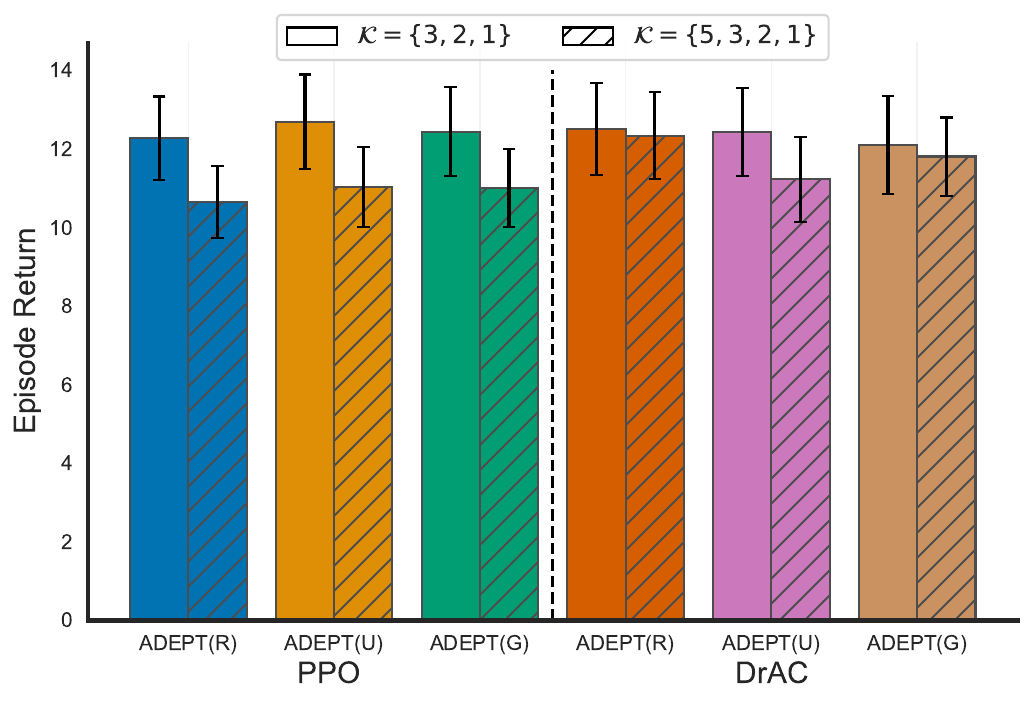}
\caption{Aggregated training performance comparison of ADEPT with different sets of NUE values. We use $c=5.0,W=10$ for ADEPT(U) and $\eta=1.0,W=10$ for ADEPT(G). The mean and standard deviation are computed across all the environments.}
\label{fig:pg_nue_ablations}
\end{center}
\vskip -0.3in
\end{figure}

% \begin{figure}[h!]
% % \vskip 0.2in
% \begin{center}
% \centerline{\includegraphics[width=\linewidth]{figures/pg_ppo_ucb_ablations.pdf}}
% \caption{Aggregated training performance comparison of PPO+ADEPT(U) with different exploration coefficients and sizes of the sliding window. The mean and standard deviation are computed across all the environments.}
% \label{fig:pg_ppo_ucb_ablations}
% \end{center}
% \vskip -0.2in
% \end{figure}

\begin{figure*}[t!]
% \vskip 0.1in
\begin{center}
\centerline{\includegraphics[width=\linewidth]{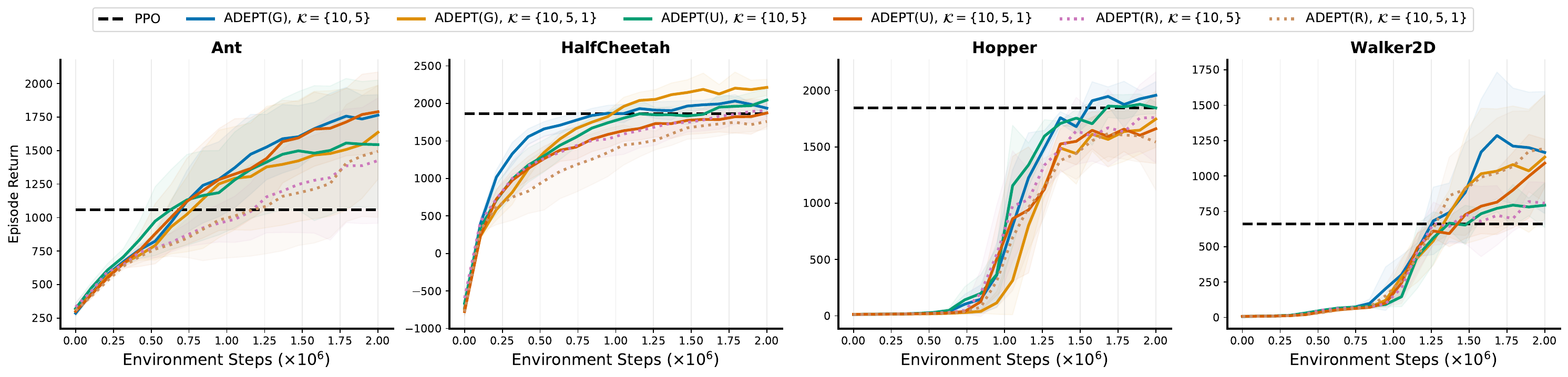}}
\caption{Performance of the PPO and its combinations with three ADEPT algorithms on the PyBullet benchmark. The mean and standard deviation are computed using five random seeds.}
\label{fig:pb_ppo}
\end{center}
\vskip -0.3in
\end{figure*}

\textbf{Generalization performance on Procgen}. We further evaluate the generalization performance of PPO, DrAC, and their combinations with ADEPT on the Procgen benchmark. Figure~\ref{fig:pg_test_scores} illustrates four aggregated evaluation metrics on the whole Procgen benchmark, in which all the scores are normalized by the mean score of PPO. For the PPO agent, all three ADEPT algorithms obtain significant performance gains regarding the four metrics. ADEPT(U) achieves a higher mean score due to its relatively aggressive scheduling strategy, while ADEPT(G) improves the IQM score by exploring more candidates. For the DrAC agent, ADEPT(R) also outperforms the vanilla DrAC agent overall, and ADEPT(U) and ADEPT(G) can still obtain remarkable performance gains. These results demonstrate that ADEPT can effectively enhance RL agents' generalization through automatic and precise learning scheduling.

\textbf{Ablation studies}. We also conducted a number of ablation experiments to study the importance of hyperparameters used in ADEPT, and the results are provided in Appendix~\ref{appendix:ablation hp}. It indicates that $c=5.0$ and $W=10$ are the overall best for ADEPT(U) in both PPO and DrAC experiments. For PPO+ADEPT(G), the best options are $\eta=1.0$ and $w=50$, while for DrAC+ADEPT(G), they are $\eta=0.1$ and $w=50$. Additionally, we examine ADEPT with different sets
of NUE values, as shown in Figure~\ref{fig:pg_nue_ablations} and Appendix~\ref{appendix:ablation nue}. The results demonstrate that a bigger $\mathcal{K}$ can only enhance ADEPT in a few environments, but degrade the overall performance on the Procgen benchmark. 

% Figure~\ref{fig:pg_ppo_ucb_ablations} illustrates the learning curves of PPO+ADEPT(U) with different $c$ and $W$, which indicates that $c=5.0$ and $W=10$ are the overall best hyperparameters. Similarly, we perform a hyperparameter search over the step size $\eta$ and the sliding window length for PPO+ADEPT(G), as illustrated in Figure~\ref{fig:pg_ppo_ts_ablations}, which indicates that $\eta=$ and $w=$ are the overall best options. For the DAAC experiments, we do a similar hyperparameter search for the two ADEPT algorithms, and the detailed comparison can be found in the Appendix~\ref{appendix:ablation}.

\begin{figure}[h!]
% \vskip 0.1in
\begin{center}
\centerline{\includegraphics[width=\linewidth]{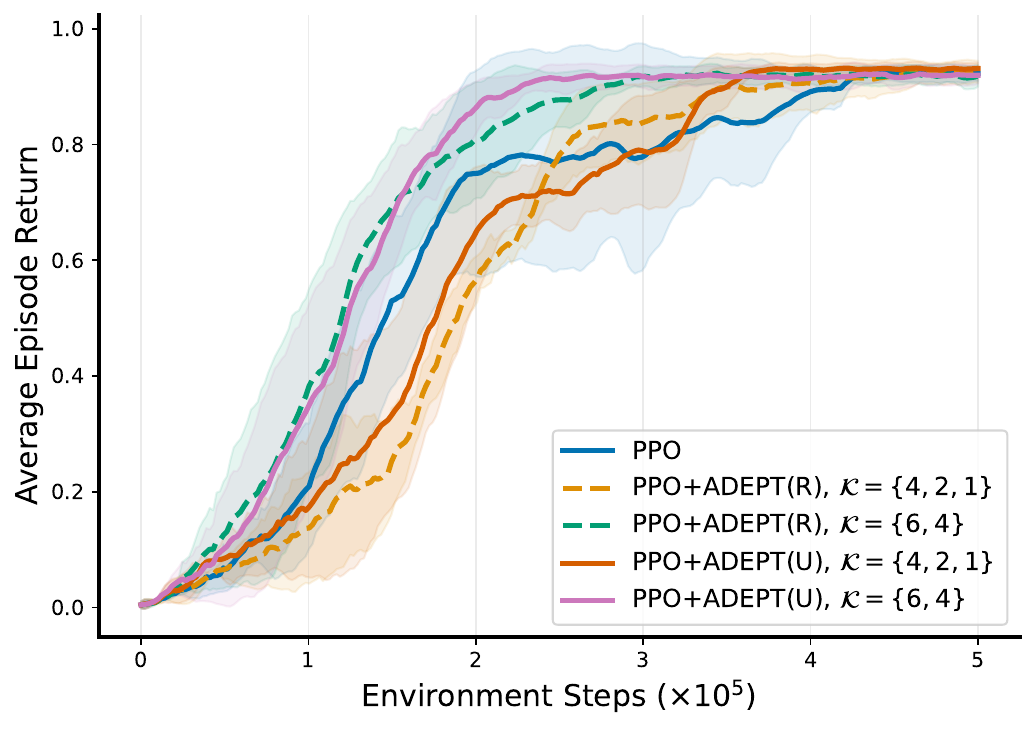}}
\caption{Aggregated performance of ADEPT(U) and ADEPT(R) on the MiniGrid benchmark. The mean and standard deviation are computed across all the environments. }
\label{fig:mg_ppo_agg}
\end{center}
\vskip -0.2in
\end{figure}

\textbf{Data efficiency on MiniGrid}. Additionally, we evaluate ADEPT on the MiniGrid benchmark with sparse-rewards and goal-oriented environments. Specifically, we conduct experiments using \textit{DoorKey-6$\times$6}, \textit{LavaGapS7}, and \textit{Empty-16$\times$16}. Figure~\ref{fig:mg_ppo_agg} illustrates the aggregated learning curves of the vanilla PPO agent, PPO+ADEPT(R), and PPO+ADEPT(U) using various sets of NUE candidates. It is obvious that ADEPT takes fewer environment steps to solve the tasks, highlighting its capability to accelerate RL algorithms in both dense and sparse-reward settings. More experimental details are provided in Appendix~\ref{appendix:exp setup}.

\textbf{Performance on continuous control tasks}. Finally, we evaluate ADEPT on the PyBullet benchmark with continuous control tasks. Four environments are utilized, namely \textit{Ant}, \textit{HalfCheetah}, \textit{Hopper}, and \textit{Walker2D}. Figure~\ref{fig:pb_ppo} illustrates the aggregated learning curves of the vanilla PPO agent and its combinations with three ADEPT algorithms. Here, we also run a hyperparameter search as the Procgen experiments and report the best results. As shown in Figure~\ref{fig:pb_ppo}, ADEPT outperforms the PPO agent with fixed NUE values, especially in \textit{Ant} and \textit{Walker2D} environments. These results underscore the effectiveness of ADEPT in enhancing RL algorithms across both discrete and continuous control tasks. Additional experimental details can be found in Appendix~\ref{appendix:exp setup}.

% \begin{figure*}[t!]
% \vskip 0.2in
% \begin{center}
% \centerline{\includegraphics[width=\linewidth]{figures/pg_ppo_ucb_decision_8.pdf}}
% \caption{The cumulative number of times ADEPT(U) selects each $K$ over the training progress.}
% \label{fig:pg_ppo_ucb_decision_8}
% \end{center}
% \vskip -0.2in
% \end{figure*}

\section{Discussion}
In this paper, we investigated the problem of improving data efficiency and generalization in deep RL and proposed a novel framework entitled ADEPT. By adaptively managing the data utilization across different learning stages, ADEPT can optimize data efficiency and significantly reduce computational overhead. In addition, ADEPT substantially enhances generalization in procedurally-generated environments. We evaluate ADEPT on Procgen, MiniGrid, and PyBullet benchmarks. Extensive simulation results demonstrate that ADEPT can effectively enhance RL algorithms with simple architecture, providing a practical solution to data-efficiency RL.

Still, there are currently remaining limitations to this work. Specifically, the decision-making process in ADEPT relies on the task return predicted by the value network, and inaccurate predictions will directly affect the scheduling quality. Furthermore, oscillatory scheduling may lead to underfitting in the value network, potentially degrading overall performance. Additionally, in the case of ADEPT(G), we assume the arm reward follows a normal distribution, which may not generalize well to all scenarios. Future work will focus on mitigating these issues by improving the robustness of value network predictions and exploring more generalized reward modeling techniques, further solidifying the applicability and reliability of ADEPT.

% \section*{Software and Data}

% If a paper is accepted, we strongly encourage the publication of software and data with the
% camera-ready version of the paper whenever appropriate. This can be
% done by including a URL in the camera-ready copy. However, \textbf{do not}
% include URLs that reveal your institution or identity in your
% submission for review. Instead, provide an anonymous URL or upload
% the material as ``Supplementary Material'' into the OpenReview reviewing
% system. Note that reviewers are not required to look at this material
% when writing their review.

\clearpage\newpage

% \section*{Acknowledgements}

% \textbf{Do not} include acknowledgements in the initial version of
% the paper submitted for blind review.

% If a paper is accepted, the final camera-ready version can (and
% usually should) include acknowledgements.  Such acknowledgements
% should be placed at the end of the section, in an unnumbered section
% that does not count towards the paper page limit. Typically, this will 
% include thanks to reviewers who gave useful comments, to colleagues 
% who contributed to the ideas, and to funding agencies and corporate 
% sponsors that provided financial support.

\section*{Impact Statement}
This paper introduces the ADEPT framework, which aims to advance deep reinforcement learning (RL) by improving data efficiency, enhancing generalization, and minimizing computational overhead. While AI has achieved remarkable success across various domains, the training processes are resource-intensive, consuming large amounts of electricity annually. This substantial energy demand contributes to increased carbon emissions, further exacerbating climate change, and results in higher operational costs for institutions involved in AI research and development. ADEPT provides a practical and scalable solution to data-efficient and computation-efficient RL, with the potential to reduce energy consumption and carbon emissions. By addressing these issues, ADEPT contributes to energy conservation and supports global sustainability efforts, promoting environmental protection while advancing AI capabilities.

% Authors are \textbf{required} to include a statement of the potential 
% broader impact of their work, including its ethical aspects and future 
% societal consequences. This statement should be in an unnumbered 
% section at the end of the paper (co-located with Acknowledgements -- 
% the two may appear in either order, but both must be before References), 
% and does not count toward the paper page limit. In many cases, where 
% the ethical impacts and expected societal implications are those that 
% are well established when advancing the field of Machine Learning, 
% substantial discussion is not required, and a simple statement such 
% as the following will suffice:

% ``This paper presents work whose goal is to advance the field of 
% Machine Learning. There are many potential societal consequences 
% of our work, none which we feel must be specifically highlighted here.''

% The above statement can be used verbatim in such cases, but we 
% encourage authors to think about whether there is content which does 
% warrant further discussion, as this statement will be apparent if the 
% paper is later flagged for ethics review.


\bibliography{example_paper}
\bibliographystyle{icml2025}

%%%%%%%%%%%%%%%%%%%%%%%%%%%%%%%%%%%%%%%%%%%%%%%%%%%%%%%%%%%%%%%%%%%%%%%%%%%%%%%
%%%%%%%%%%%%%%%%%%%%%%%%%%%%%%%%%%%%%%%%%%%%%%%%%%%%%%%%%%%%%%%%%%%%%%%%%%%%%%%
% APPENDIX
%%%%%%%%%%%%%%%%%%%%%%%%%%%%%%%%%%%%%%%%%%%%%%%%%%%%%%%%%%%%%%%%%%%%%%%%%%%%%%%
%%%%%%%%%%%%%%%%%%%%%%%%%%%%%%%%%%%%%%%%%%%%%%%%%%%%%%%%%%%%%%%%%%%%%%%%%%%%%%%
\newpage
\appendix
\onecolumn
\section{Algorithmic Baselines}\label{appendix:baseline}
% \subsection{A2C}
% Advantage actor-critic (A2C) \cite{mnih2016asynchronous} is a synchronous version of the asynchronous advantage actor-critic (A3C) algorithm that stabilizes the learning process by training multiple agents in parallel and aggregating their gradients. A2C leverages a policy network $\pi_{\bm\theta}(\bm{a}|\bm{s})$ and a value network $V_{\bm\phi}(\bm{s})$ to guide action and estimate the expected return, respectively.

% The loss function for policy updates is defined as:
% \begin{equation}
%     L_{\pi}(\bm{\theta})=-\mathbb{E}_{\tau\sim\pi}\left[A^{\pi}(\bm{a}|\bm{s})\log \pi_{\bm\theta}(\bm{a}|\bm{s})\right],
% \end{equation}
% where $\tau$ is the sampled trajectory and $A^{\pi}(\bm{a}|\bm{s})$ is the advantage function.

% The value network is trained to minimize the error between the predicted return and a target of discounted returns computed with generalized advantage estimation (GAE) \cite{schulman2015high}:
% \begin{equation}
%     L_{V}(\bm{\phi})=\mathbb{E}_{\tau\sim\pi}\left[\left(V_{\bm\phi}(\bm{s})-V_{t}^{\rm target}\right)^{2}\right]
% \end{equation}

\subsection{PPO}
Proximal policy optimization (PPO) \cite{schulman2017proximal} is an on-policy algorithm that is designed to improve the stability and sample efficiency of policy gradient methods, which uses a clipped surrogate objective function to avoid large policy updates. 

The policy loss is defined as:
\begin{equation}
    L_{\pi}(\bm{\theta})=-\mathbb{E}_{\tau\sim\pi}\left[\min\left(\rho_{t}(\bm{\theta})A_{t},{\rm clip}\left(\rho_{t}(\bm{\theta}),1-\epsilon,1+\epsilon\right)A_{t}\right)\right],
\end{equation}
where 
\begin{equation}
    \rho_{t}(\bm{\theta})=\frac{\pi_{\bm\theta}(\bm{a}_{t}|\bm{s}_{t})}{\pi_{\bm\theta_{\rm old}}(\bm{a}_{t}|\bm{s}_{t})},
\end{equation}
and $\epsilon$ is a clipping range coefficient.

Meanwhile, the value network is trained to minimize the error between the predicted return and a target of discounted returns computed with generalized advantage estimation (GAE) \cite{schulman2015high}:
\begin{equation}
    L_{V}(\bm{\phi})=\mathbb{E}_{\tau\sim\pi}\left[\left(V_{\bm\phi}(\bm{s})-V_{t}^{\rm target}\right)^{2}\right].
\end{equation}

\subsection{DrAC}
Data-regularized actor-critic (DrAC) \cite{raileanu2020automatic} is proposed to address the challenge of generalization in procedurally-generated environments by introducing data augmentation during training. Moreover, DrAC utilizes two regularization terms to constrain the agent’s policy and value function to be invariant to various state transformations. 

The policy network is trained to minimize two parts of losses:
\begin{equation}
    L_{\pi}(\bm{\theta})=L_{\pi}^{\rm PPO}(\bm{\theta})+G_{\pi}(\bm{\theta}),
\end{equation}
where
\begin{equation}
    G_{\pi}(\bm{\theta})=D_{\rm KL}\left(\pi_{\rm \theta}(\bm{a}|\bm{s})\Vert\pi_{\rm \theta}(\bm{a}|f(\bm{s})\right),
\end{equation}
and $D_{\rm KL}$ is the Kullback–Leibler divergence and $f$ is a mapping that satisfies
\begin{equation}
    V_{\bm\phi}(\bm{s})=V_{\bm\phi}(f(\bm{s})), \pi_{\rm \theta}(\bm{a}|\bm{s})=\pi_{\rm \theta}(\bm{a}|f(\bm{s})).
\end{equation}

Similarly, the value network is also trained using two parts of losses:
\begin{equation}
    L_{V}=L_{V}^{\rm PPO}(\bm{\phi})+\left[V_{\bm\phi}(\bm{s})-V_{\bm\phi}(f(\bm{s}))\right]^{2}.
\end{equation}

% \subsection{DAAC}
% Decoupled advantage actor-critic (DAAC) \cite{raileanu2021decoupling} is proposed to address the challenge of generalization in complex environments. It is designed to learn a policy and a value function separately, which is different from the standard practice of sharing parameters (e.g., the encoder for extracting features from observations) between the policy and value networks in actor-critic methods. 

% The policy network is trained to minimize two parts of losses:
% \begin{equation}
% \begin{aligned}
%     L_{\pi}(\bm{\theta})&=L_{\pi}^{\rm PPO}+L_{A}(\bm{\theta}),\\
%     L_{A}(\bm{\theta})&=\mathbb{E}_{\tau\sim\pi}\left[\left(A_{\rm\theta}(\bm{a}_{t}|\bm{s}_{t})-\hat{A}_{t}\right)\right],
% \end{aligned}
% \end{equation}
% where $A_{\rm\theta}$ the the predicted advantage by the policy network, and $\hat{A}_{t}$ is is the corresponding generalized advantage estimate
% at time step $t$.

% DAAC also utilizes the same loss function as A2C to train the value network. Moreover, during training, DAAC alternates between $E_{\pi}$ epochs for training
% the policy network and $E_{V}$ epochs for training the value
% network every $N_{\pi}$ policy update.

\clearpage\newpage

\section{Experimental Setup}\label{appendix:exp setup}

\subsection{Procgen}
\begin{figure*}[h!]
\vskip 0.2in
\begin{center}
\centerline{\includegraphics[width=\linewidth]{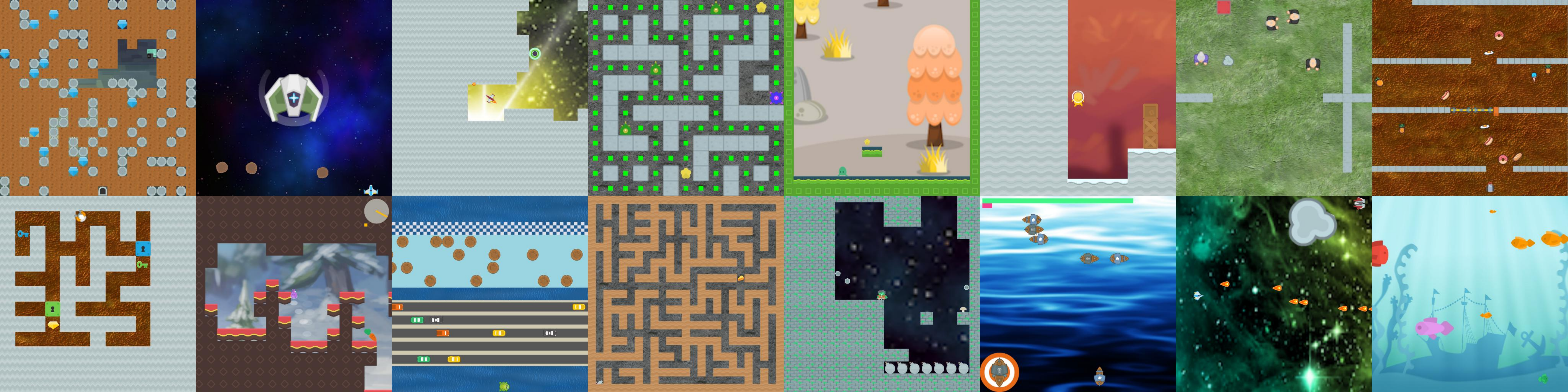}}
\caption{Screenshots of the sixteen Procgen environments.}
\label{fig:procgen_screenshots}
\end{center}
\vskip -0.2in
\end{figure*}

\textbf{PPO+ADEPT}. In this part, we leverage the implementation of CleanRL \cite{huang2022cleanrl} for the PPO algorithm. Table~\ref{tb:ppo_daac_hp} illustrates the PPO hyperparameters, which remain fixed throughout all the experiments.

Since the reported overall best $K$ in \cite{cobbe2020leveraging} is 3, the candidates of NUE are set as $\mathcal{K}=\{3,2,1\}$ for all the experiments. For \textbf{ADEPT(U)}, we ran a hyperparameter search over the exploration coefficient $c\in\{0.1, 1.0, 5.0\}$ and the size of the sliding window used to compute the $Q$-values $W\in[10, 50, 100]$. We found that $c=5.0, W=10$ are the best hyperparameters overall. Similarly, for \textbf{ADEPT(G)}, we ran a hyperparameter search over the learning rate $\eta\in\{0.1, 0.5, 1.0\}$ and the size of the sliding window used to compute the $Q$-values $W\in[10, 50, 100]$, and $\alpha=1.0, W=50$ are the overall best hyperparameters. These values are used to obtain the results reported in the paper.

\begin{table}[h!]
\centering
\caption{The shared hyperparameters for PPO and DrAC on Procgen. These remain fixed for all experiments.}
\label{tb:ppo_daac_hp}
\vskip 0.1in
\begin{tabular}{ll}
\toprule[1.0pt]
\textbf{Hyperparameter}             & \textbf{Value}    \\ \midrule[1.0pt]
Observation downsampling   & (64, 64) \\
Observation normalization  & / 255.   \\
Reward normalization       & Yes for PPO, No for DrAC     \\
LSTM                       & No       \\
Stacked frames             & No       \\
Environment steps          & 25000000 \\
Episode steps              & 256      \\
Number of workers          & 1        \\
Environments per worker    & 64       \\
Optimizer                  & Adam     \\
Learning rate              & 5e-4     \\
GAE coefficient            & 0.95     \\
Action entropy coefficient & 0.01     \\
Value loss coefficient     & 0.5      \\
Value clip range           & 0.2      \\
Max gradient norm          & 0.5      \\
Batch size                 & 2048     \\
Discount factor            & 0.999    \\ \bottomrule[1.0pt]
\end{tabular}
\vskip -0.1in
\end{table}

\textbf{DrAC+ADEPT}. In this part, we use the official implementation \cite{raileanu2020automatic} of DrAC for the experiments, and Table~\ref{tb:ppo_daac_hp} lists the shared and fixed hyperparameters. For each Procgen environment, Table~\ref{tb:best_aug_drac} lists the best augmentation method of DrAC as reported in \cite{raileanu2021decoupling}. The candidates of NUE are also set as $\mathcal{K}=\{3,2,1\}$ for all the experiments. For \textbf{ADEPT(U)} and \textbf{ADEPT(G)}, we run the same hyperparameter search as the experiments of PPO+ADEPT and report the best results.

\begin{table}[h!]
\centering
\caption{Best augmentation type of DrAC for each Procgen environment.}
\label{tb:best_aug_drac}
\vskip 0.1in
\begin{tabular}{lllllllll}
\toprule[1.0pt]
\textbf{Env.} & BigFish     & StarPilot & FruitBot     & BossFight & Ninja        & Plunder & CaveFlyer & CoinRun      \\ \midrule[1.0pt]
\textbf{Aug.}  & Crop        & Crop      & Crop         & Flip      & Color-jitter & Crop    & Rotate    & Random-conv  \\ \toprule[1.0pt]
\textbf{Env.} & Jumper      & Chaser    & Climber      & Dodgeball & Heist        & Leaper  & Maze      & Miner        \\ \midrule[1.0pt]
\textbf{Aug.}  & Random-conv & Crop      & Color-jitter & Crop      & Crop         & Crop    & Crop      & Color-jitter \\ \bottomrule[1.0pt]
\end{tabular}
\vskip -0.1in
\end{table}

\subsection{MiniGrid}
\begin{figure*}[h!]
\vskip 0.2in
\begin{center}
\centerline{\includegraphics[width=0.7\linewidth]{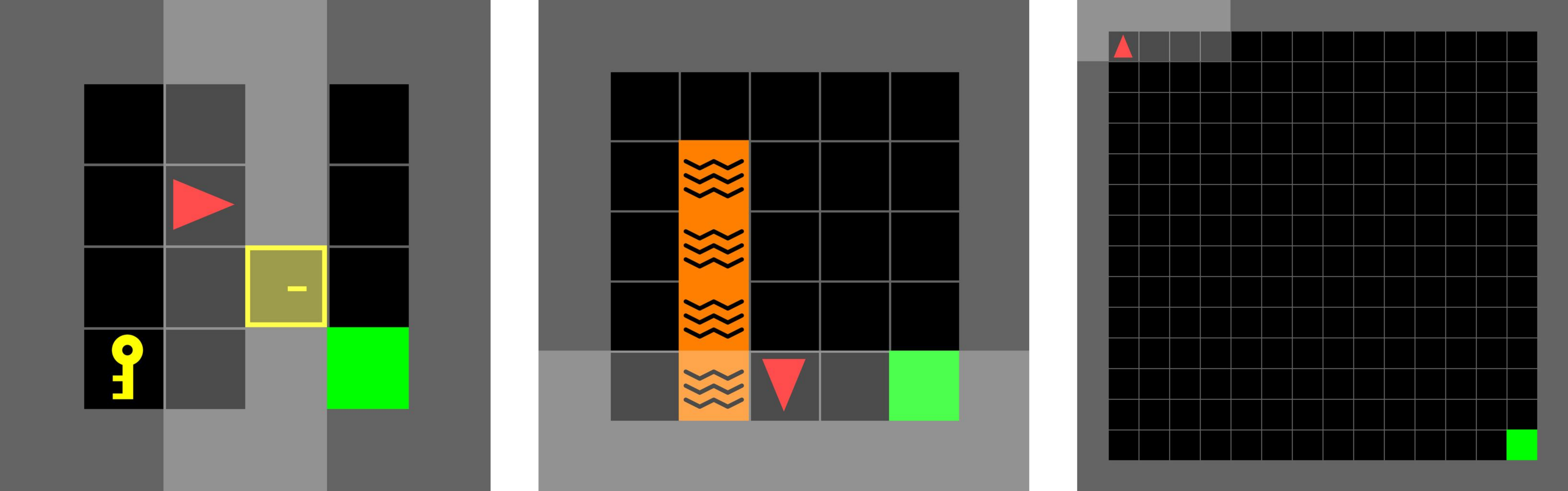}}
\caption{Screenshots of the three MiniGrid environments. From left to right: \textit{DoorKey-6$\times$6}, \textit{LavaGapS7}, and \textit{Empty-16$\times$16}.}
\label{fig:minigrid_screenshots}
\end{center}
\vskip -0.2in
\end{figure*}

In this part, we use the implementation of \cite{MinigridMiniworld23} for the PPO agent. Since the reported $K=4$, we evaluate PPO+ADEPT(R) and PPO+ADEPT(U) using three NUE sets: $\{4,2,1\}$, $\{6,4\}$, and $\{8,4\}$. For ADEPT(U), the exploration coefficient $c$ is set as $1.0$, and the size of the sliding window is set as $50$. Finally, Table~\ref{tb:mg_pb_ppo_hp} illustrates the PPO hyperparameters, which remain fixed throughout all the experiments. 

\subsection{PyBullet}
\begin{figure*}[h!]
\vskip 0.2in
\begin{center}
\centerline{\includegraphics[width=0.75\linewidth]{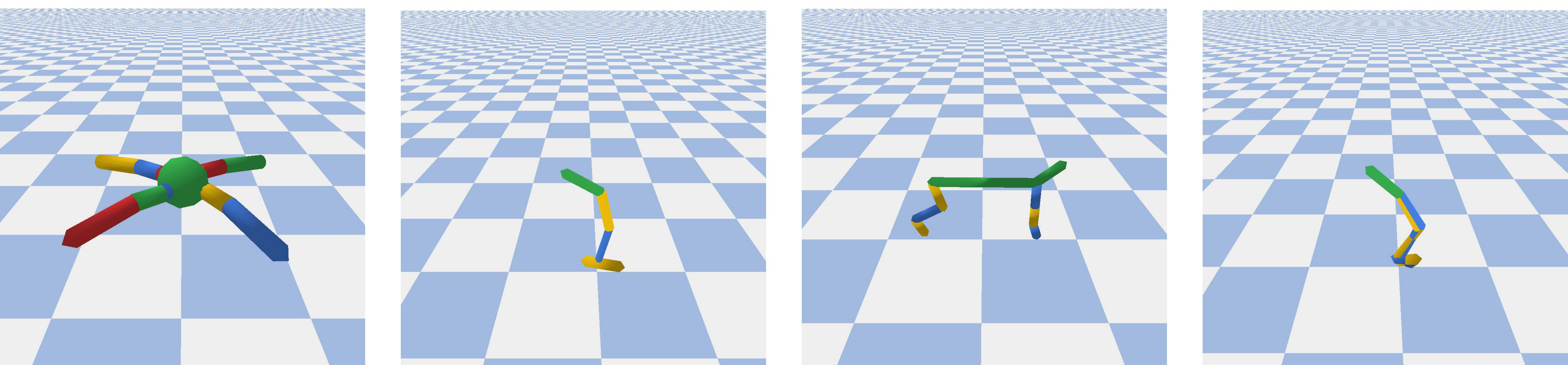}}
\caption{Screenshots of the four PyBullet environments. From left to right: \textit{Ant}, \textit{Hopper}, \textit{HalfCheetah}, and \textit{Walker2D}.}
\label{fig:pybullet_screenshots}
\end{center}
\vskip -0.2in
\end{figure*}

Finally, we perform the experiments on the PyBullet benchmark using the PPO implementation of \cite{pytorchrl}. Since \cite{raffin2022smooth} reported the best $K=10$, we set the NUE candidates as $\mathcal{K}=\{10, 5, 1\}$ and $\mathcal{K}=\{10, 5\}$. Then we test the PPO agent with three ADEPT algorithms. Moreover, we run a similar hyperparameter search as the Procgen experiments and report the best results of each method. Table~\ref{tb:mg_pb_ppo_hp} illustrates the PPO hyperparameters that remain fixed throughout all the experiments. 

\begin{table}[h!]
\centering
\caption{The shared hyperparameters for PPO on MiniGrid and PyBullet. These remain fixed for all experiments.}
\label{tb:mg_pb_ppo_hp}
\vskip 0.15in
\begin{tabular}{lll}
\toprule[1.0pt]
\textbf{Hyperparameter}             & \textbf{MiniGrid} & \textbf{PyBullet}   \\ \midrule[1.0pt]
Observation downsampling   & (7,7,3)  & N/A      \\
Observation normalization  & No       & Yes      \\
Reward normalization       & No       & Yes      \\
LSTM                       & No       & No       \\
Stacked frames             & No       & N/A      \\
Environment steps          & 500000   & 2000000  \\
Episode steps              & 128      & 2048     \\
Number of workers          & 1        & 1        \\
Environments per worker    & 16       & 1        \\
Optimizer                  & Adam     & Adam     \\
Learning rate              & 1e-3     & 2e-4     \\
GAE coefficient            & 0.95     & 0.95     \\
Action entropy coefficient & 0.01     & 0        \\
Value loss coefficient     & 0.5      & 0.5      \\
Value clip range           & 0.2      & N/A      \\
Max gradient norm          & 0.5      & 0.5      \\
Batch size                 & 256      & 64       \\
Discount factor            & 0.99     & 0.99     \\ \bottomrule[1.0pt]
\end{tabular}
\vskip -0.1in
\end{table}

%%%%%%%%%%%%%%%%%%%%%%%%%%%%%%%%%%%%%%%%%%%%%%%%%%%%%%%%%%%%%%%%%%%%%%%%%%%%%%%
%%%%%%%%%%%%%%%%%%%%%%%%%%%%%%%%%%%%%%%%%%%%%%%%%%%%%%%%%%%%%%%%%%%%%%%%%%%%%%%

\clearpage\newpage

\section{Learning Curves}\label{appendix:curves}

\subsection{PPO+ADEPT(R)+Procgen}

\begin{figure*}[h!]
\vskip 0.2in
\begin{center}
\centerline{\includegraphics[width=\linewidth]{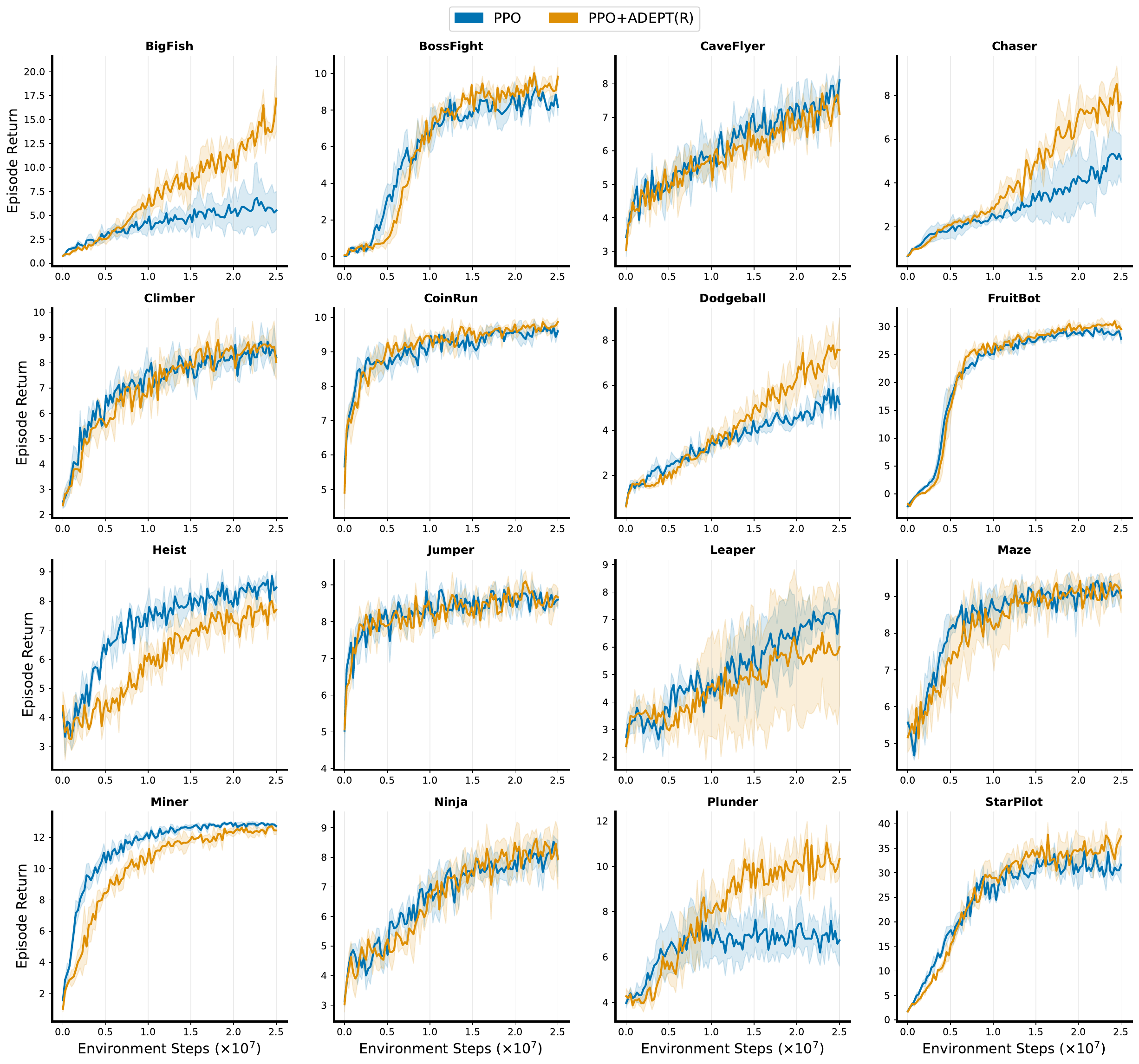}}
\caption{Learning curves of the vanilla PPO agent and PPO+ADEPT(R). The mean and standard deviation are computed over five runs with different seeds.}
\label{fig:pg_ppo_rr}
\end{center}
\vskip -0.2in
\end{figure*}

\clearpage\newpage

\subsection{PPO+ADEPT(U)+Procgen}

\begin{figure*}[h!]
\vskip 0.2in
\begin{center}
\centerline{\includegraphics[width=\linewidth]{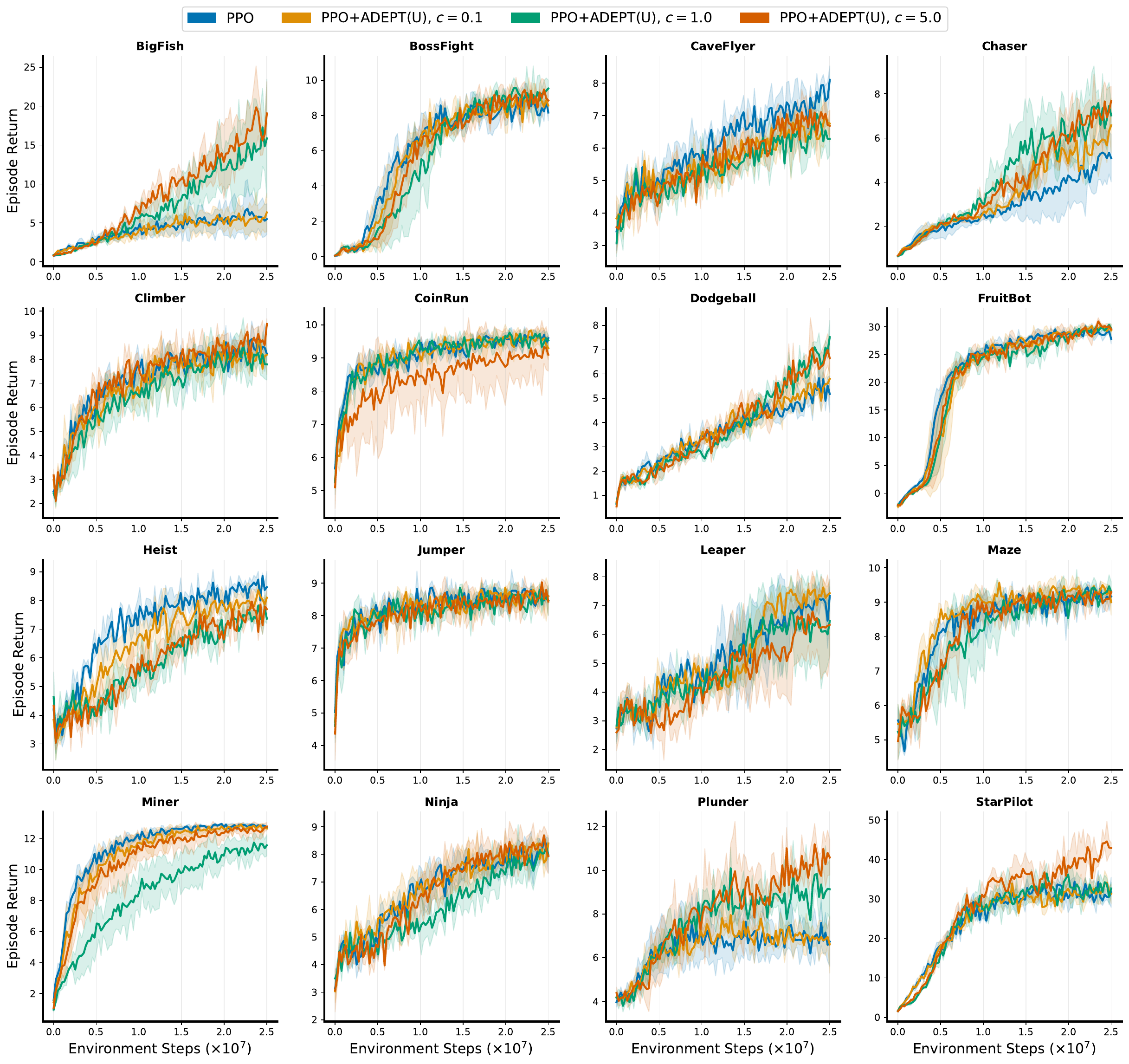}}
\caption{Learning curves of the vanilla PPO agent and PPO+ADEPT(U) with different exploration coefficients. Here, the size $W$ of the sliding window is set as 10. The mean and standard deviation are computed over five runs with different seeds.}
\label{fig:pg_ppo_ucb_expl}
\end{center}
\vskip -0.2in
\end{figure*}

\begin{figure*}[h!]
\vskip 0.2in
\begin{center}
\centerline{\includegraphics[width=\linewidth]{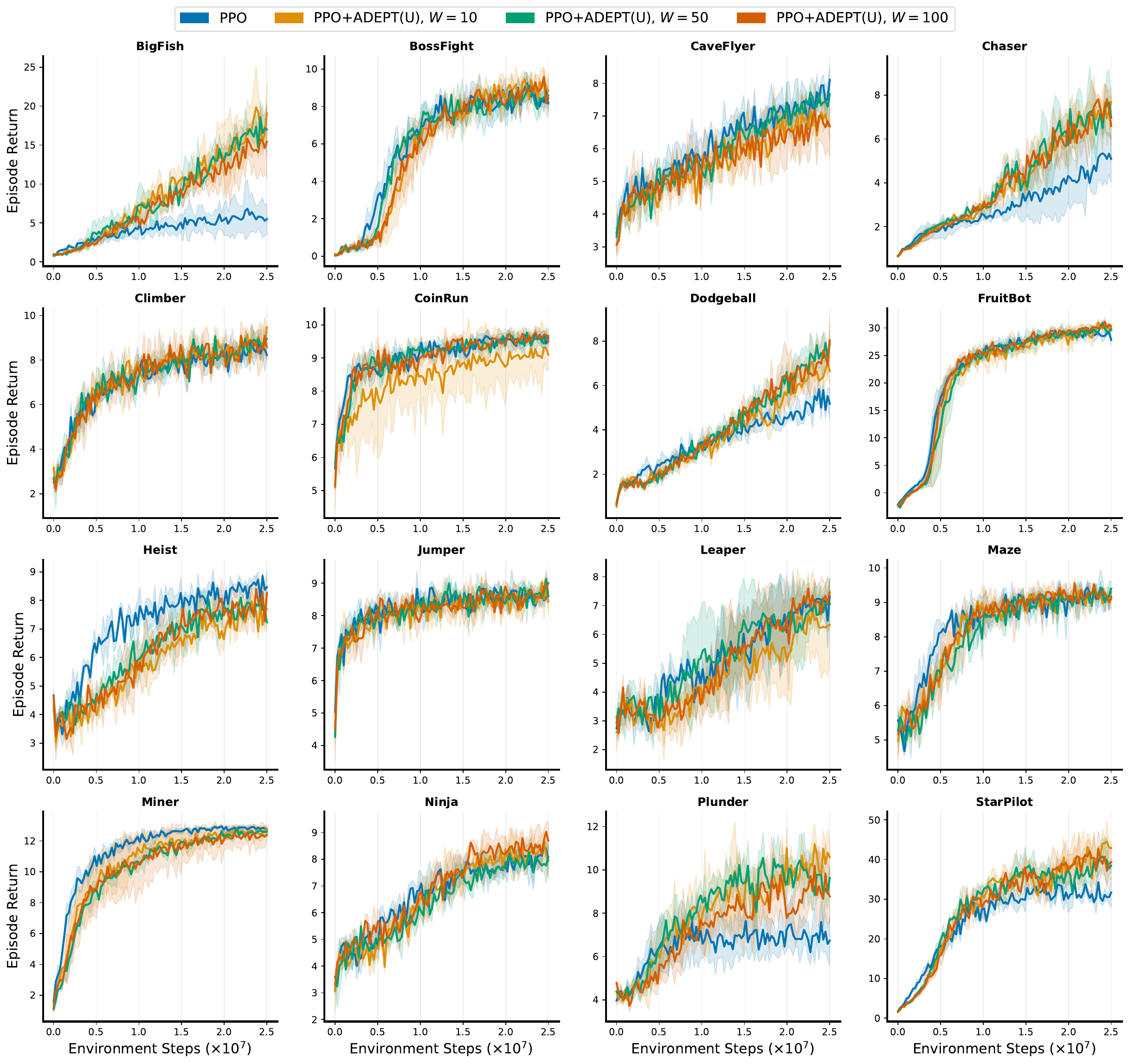}}
\caption{Learning curves of the vanilla PPO agent and PPO+ADEPT(U) with different sizes of the sliding window. Here, the exploration coefficient $c$ is set as 5.0. The mean and standard deviation are computed over five runs with different seeds.}
\label{fig:pg_ppo_ucb_window}
\end{center}
\vskip -0.2in
\end{figure*}

\clearpage\newpage

\subsection{PPO+ADEPT(G)+Procgen}

\begin{figure*}[h!]
\vskip 0.2in
\begin{center}
\centerline{\includegraphics[width=\linewidth]{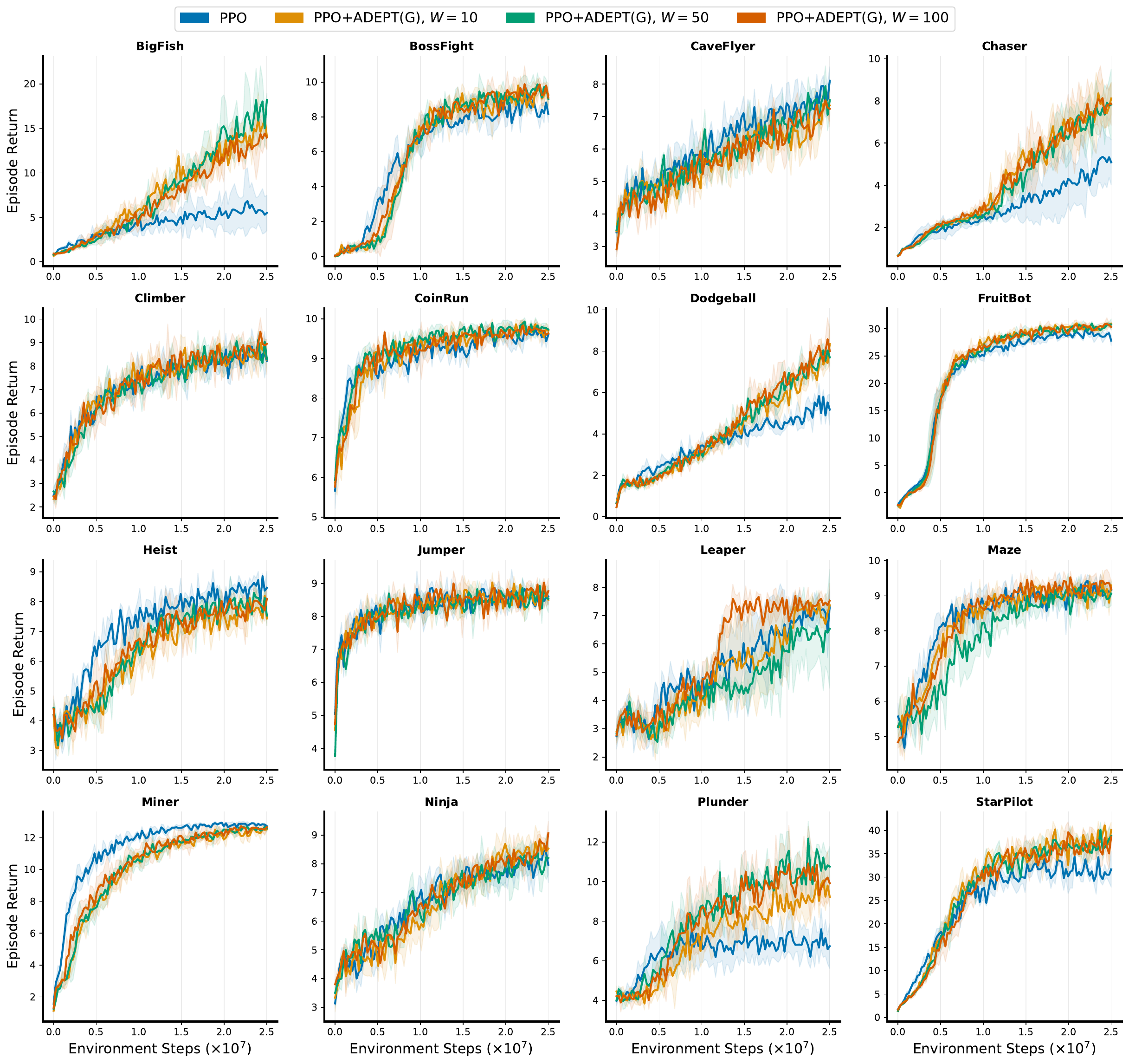}}
\caption{Learning curves of the vanilla PPO agent and PPO+ADEPT(G) with different sizes of the sliding window. Here, the learning rate $\eta$ is set as 1.0. The mean and standard deviation are computed over five runs with different seeds.}
\label{fig:pg_ppo_ts_window}
\end{center}
\vskip -0.2in
\end{figure*}

\begin{figure*}[h!]
\vskip 0.2in
\begin{center}
\centerline{\includegraphics[width=\linewidth]{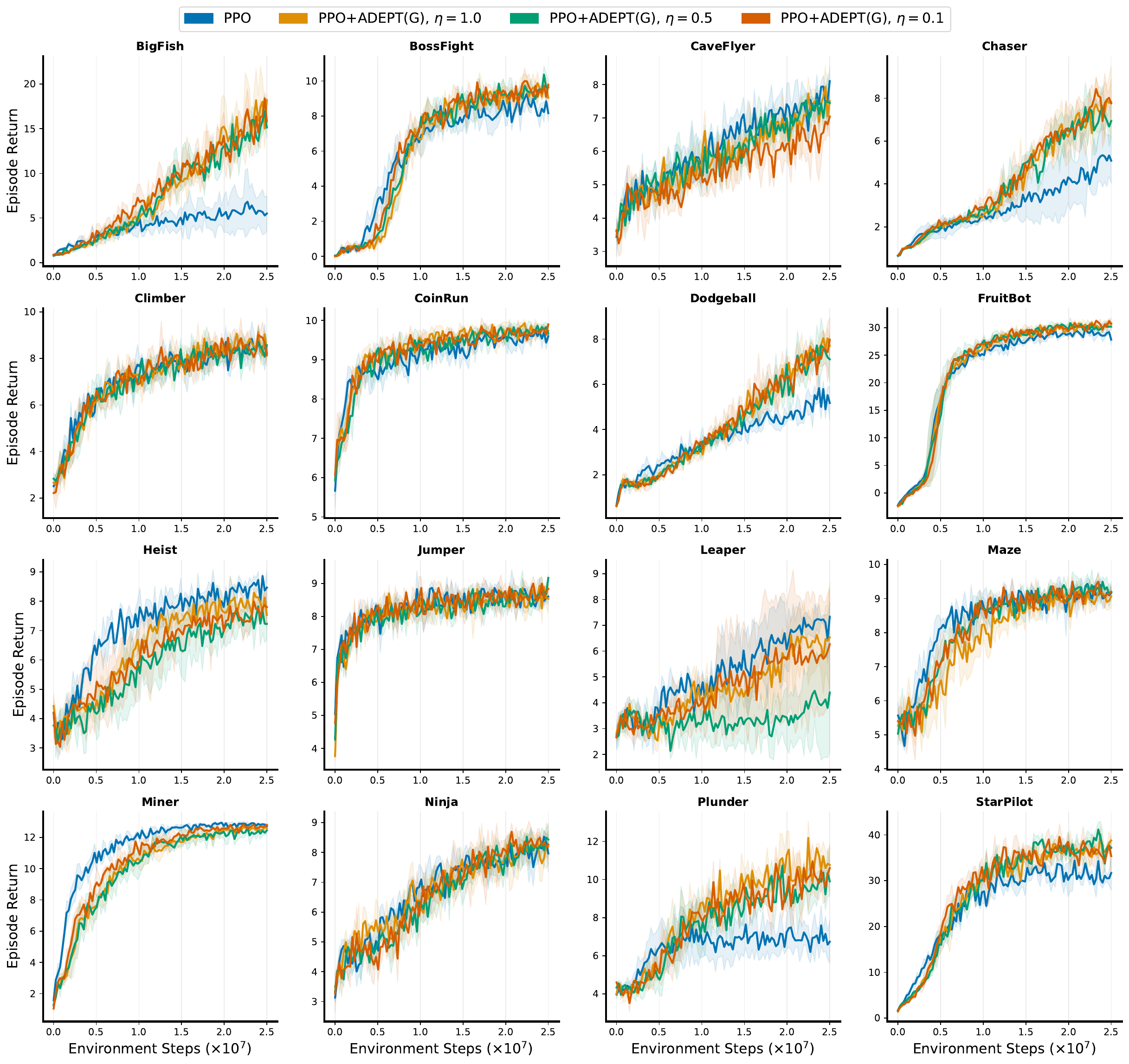}}
\caption{Learning curves of the vanilla PPO agent and PPO+ADEPT(G) with different learning rates. Here, the size $W$ of the sliding window is set as 50. The mean and standard deviation are computed over five runs with different seeds.}
\label{fig:pg_ppo_ts_eta}
\end{center}
\vskip -0.2in
\end{figure*}

\clearpage\newpage

\subsection{DrAC+ADEPT(R)+Procgen}

\begin{figure*}[h!]
\vskip 0.2in
\begin{center}
\centerline{\includegraphics[width=\linewidth]{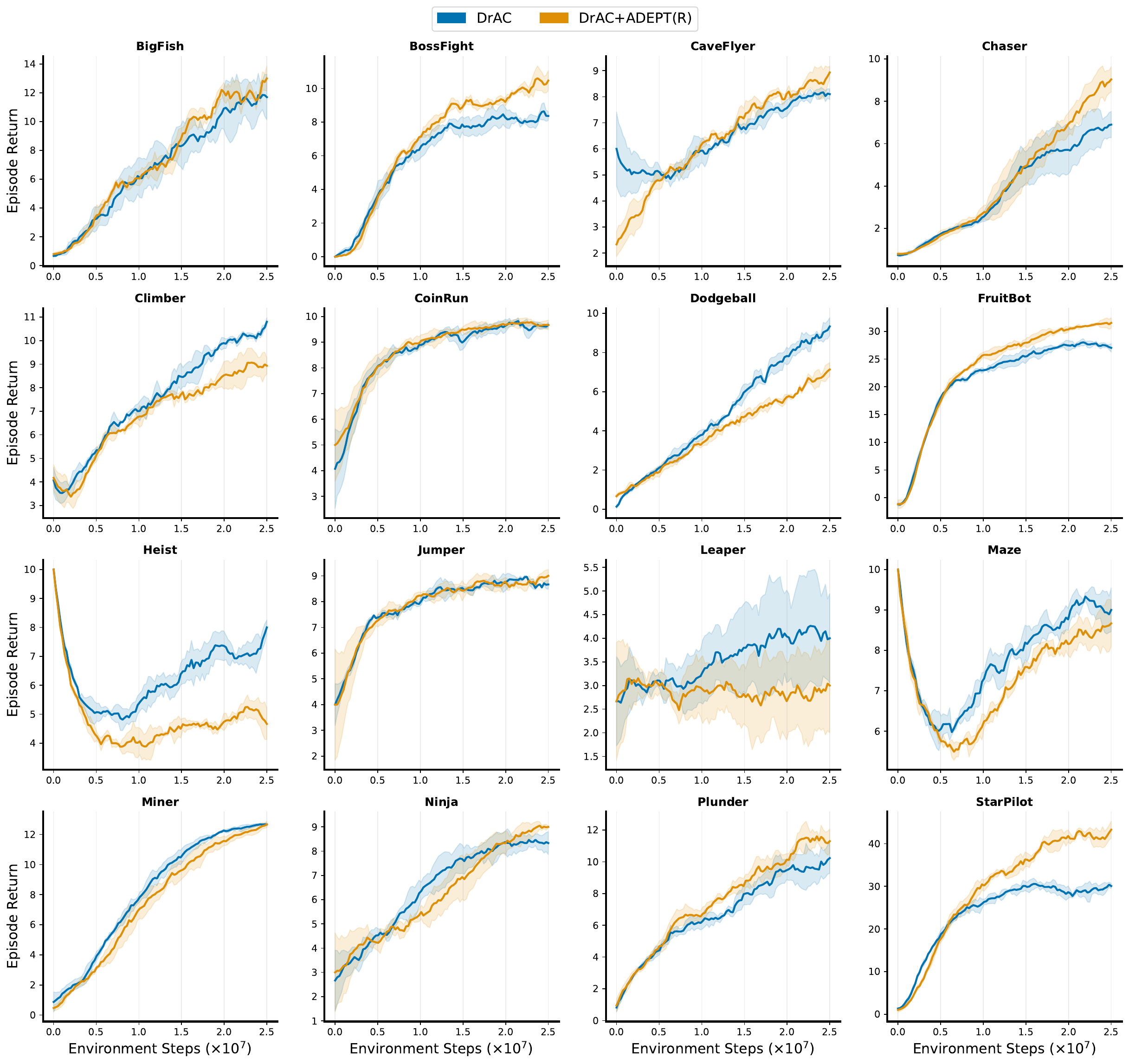}}
\caption{Learning curves of the vanilla DrAC agent and DrAC+ADEPT(R). The mean and standard deviation are computed over five runs with different seeds.}
\label{fig:pg_drac_rr}
\end{center}
\vskip -0.2in
\end{figure*}

\clearpage\newpage

\subsection{DrAC+ADEPT(U)+Procgen}

\begin{figure*}[h!]
\vskip 0.2in
\begin{center}
\centerline{\includegraphics[width=\linewidth]{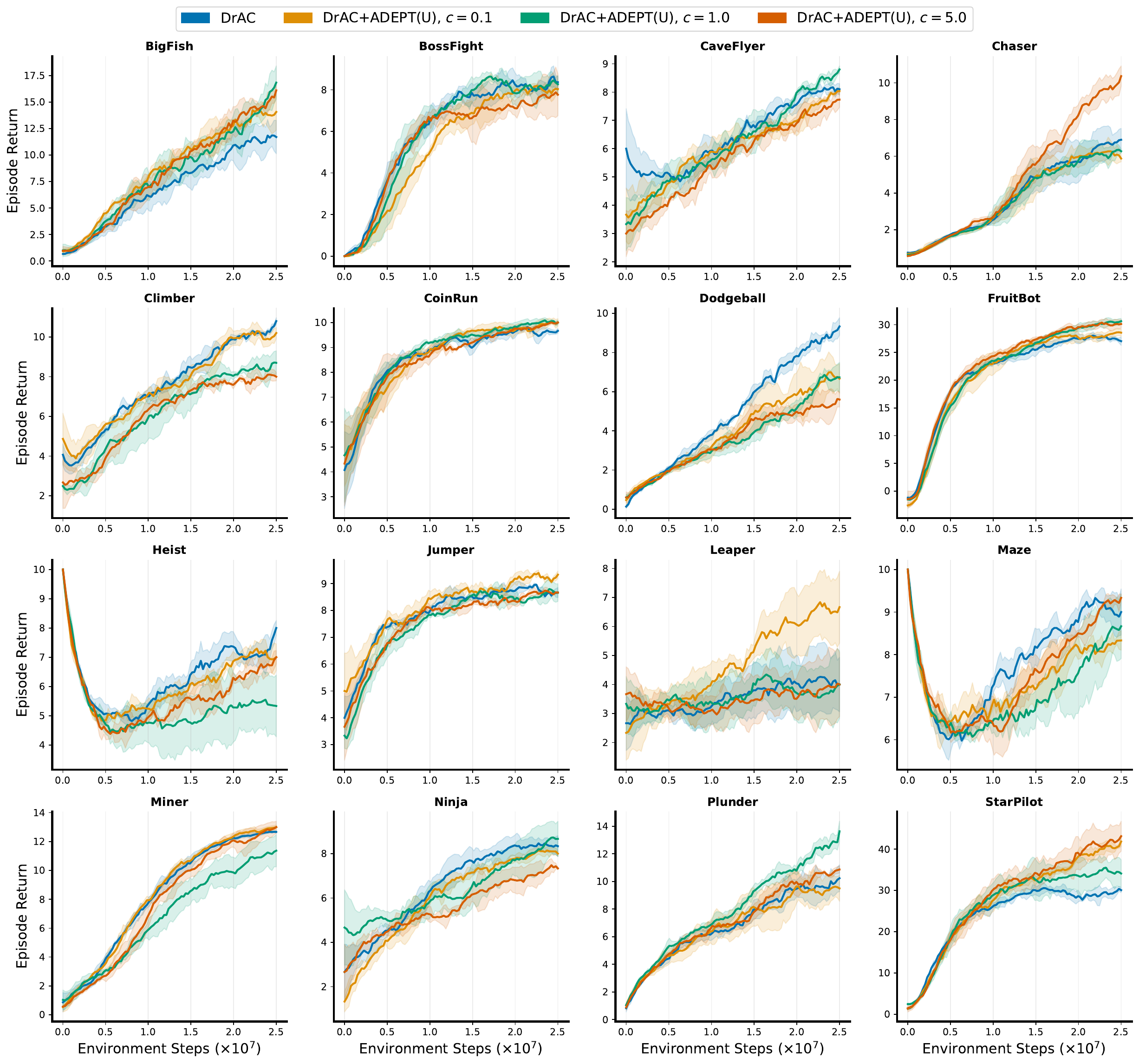}}
\caption{Learning curves of the vanilla DrAC agent and DrAC+ADEPT(U) with different exploration coefficients. Here, the size $W$ of the sliding window is set as 10. The mean and standard deviation are computed over five runs with different seeds.}
\label{fig:pg_drac_ucb_expl}
\end{center}
\vskip -0.2in
\end{figure*}

\begin{figure*}[h!]
\vskip 0.2in
\begin{center}
\centerline{\includegraphics[width=\linewidth]{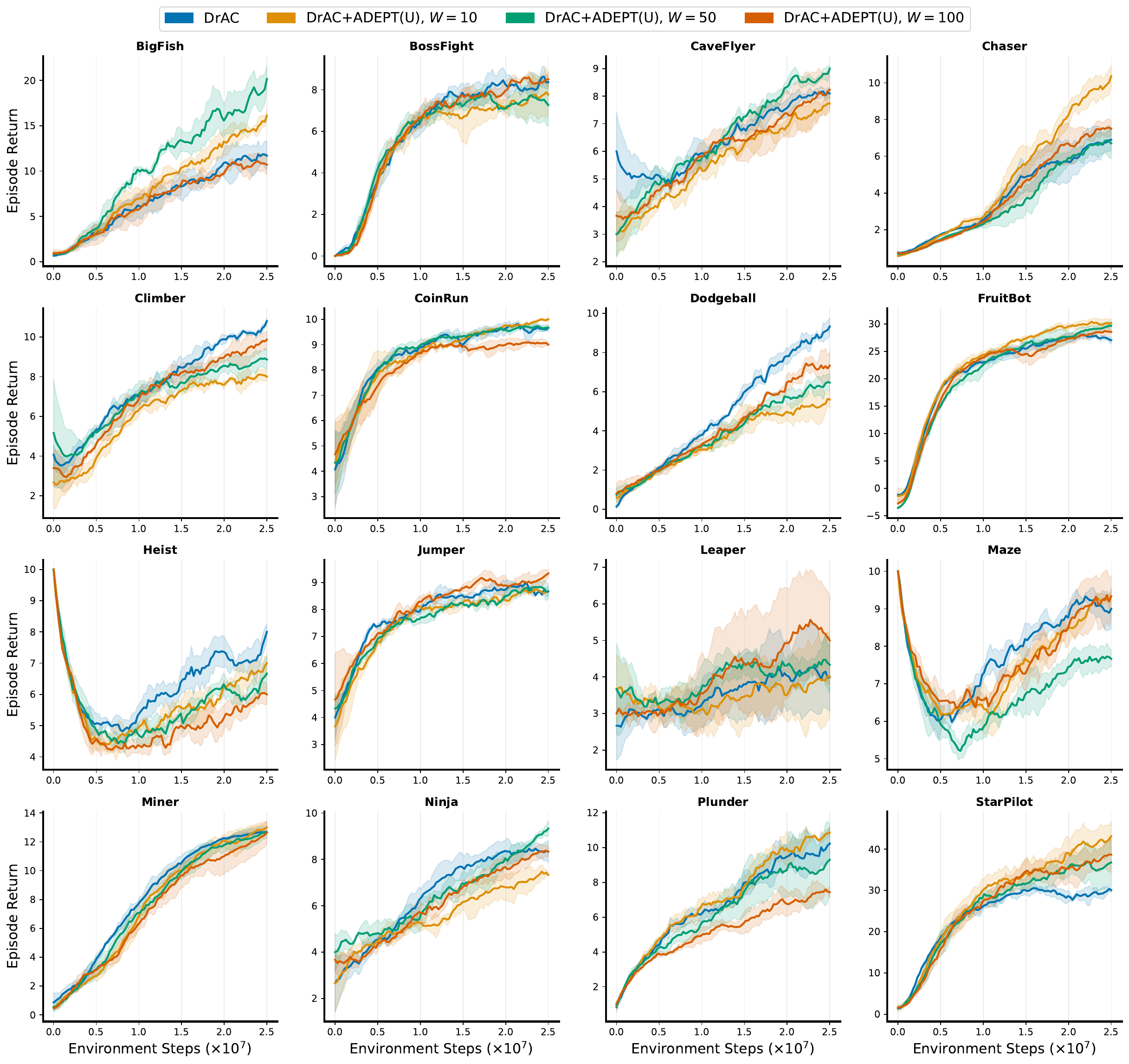}}
\caption{Learning curves of the vanilla DrAC agent and DrAC+ADEPT(U) with different sizes of the sliding window. Here, the exploration coefficient $c$ is set as 5.0. The mean and standard deviation are computed over five runs with different seeds.}
\label{fig:pg_drac_ucb_window}
\end{center}
\vskip -0.2in
\end{figure*}

\clearpage\newpage

\subsection{DrAC+ADEPT(G)+Procgen}

\begin{figure*}[h!]
\vskip 0.2in
\begin{center}
\centerline{\includegraphics[width=\linewidth]{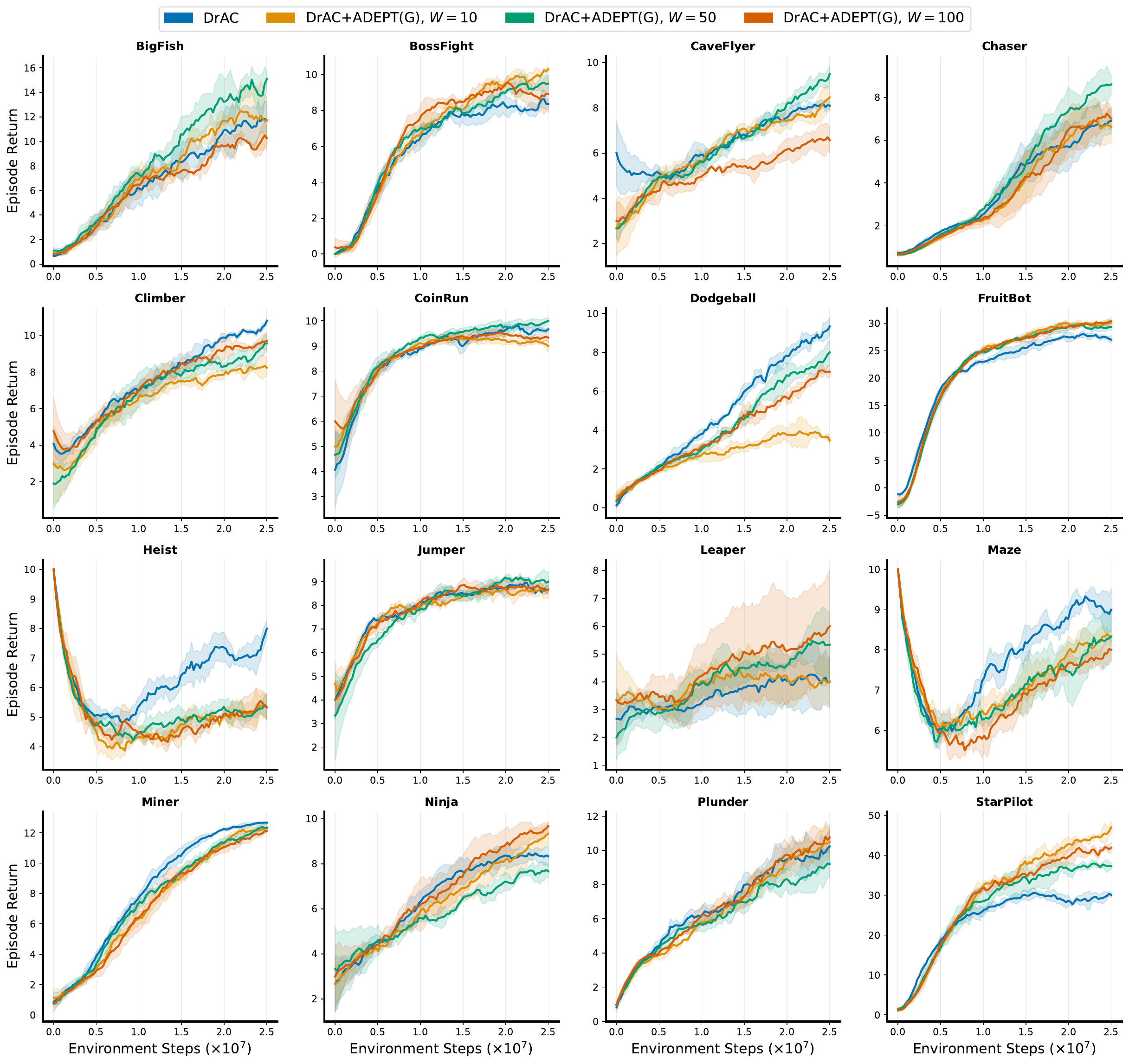}}
\caption{Learning curves of the vanilla DrAC agent and DrAC+ADEPT(G) with different sizes of the sliding window. Here, the learning rate $\eta$ is set as 1.0. The mean and standard deviation are computed over five runs with different seeds.}
\label{fig:pg_drac_ts_window}
\end{center}
\vskip -0.2in
\end{figure*}

\begin{figure*}[h!]
\vskip 0.2in
\begin{center}
\centerline{\includegraphics[width=\linewidth]{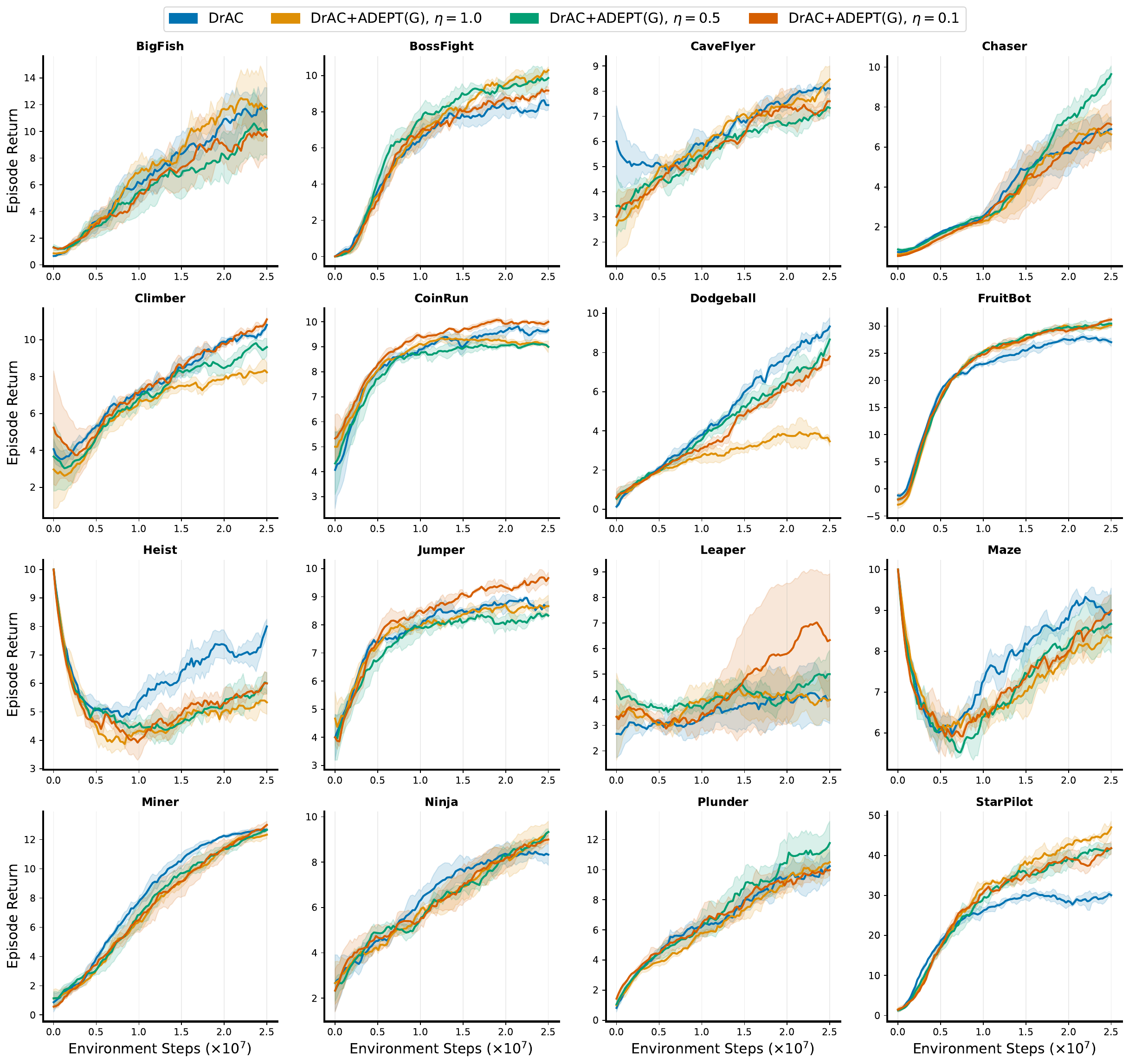}}
\caption{Learning curves of the vanilla DrAC agent and DrAC+ADEPT(G) with different learning rates. Here, the size $W$ of the sliding window is set as 50. The mean and standard deviation are computed over five runs with different seeds.}
\label{fig:pg_drac_ts_eta}
\end{center}
\vskip -0.2in
\end{figure*}

\clearpage\newpage

\section{Data Efficiency Comparison}\label{appendix:efficiency}
\begin{figure*}[h!]
\vskip 0.2in
\centering
\includegraphics[width=\linewidth]{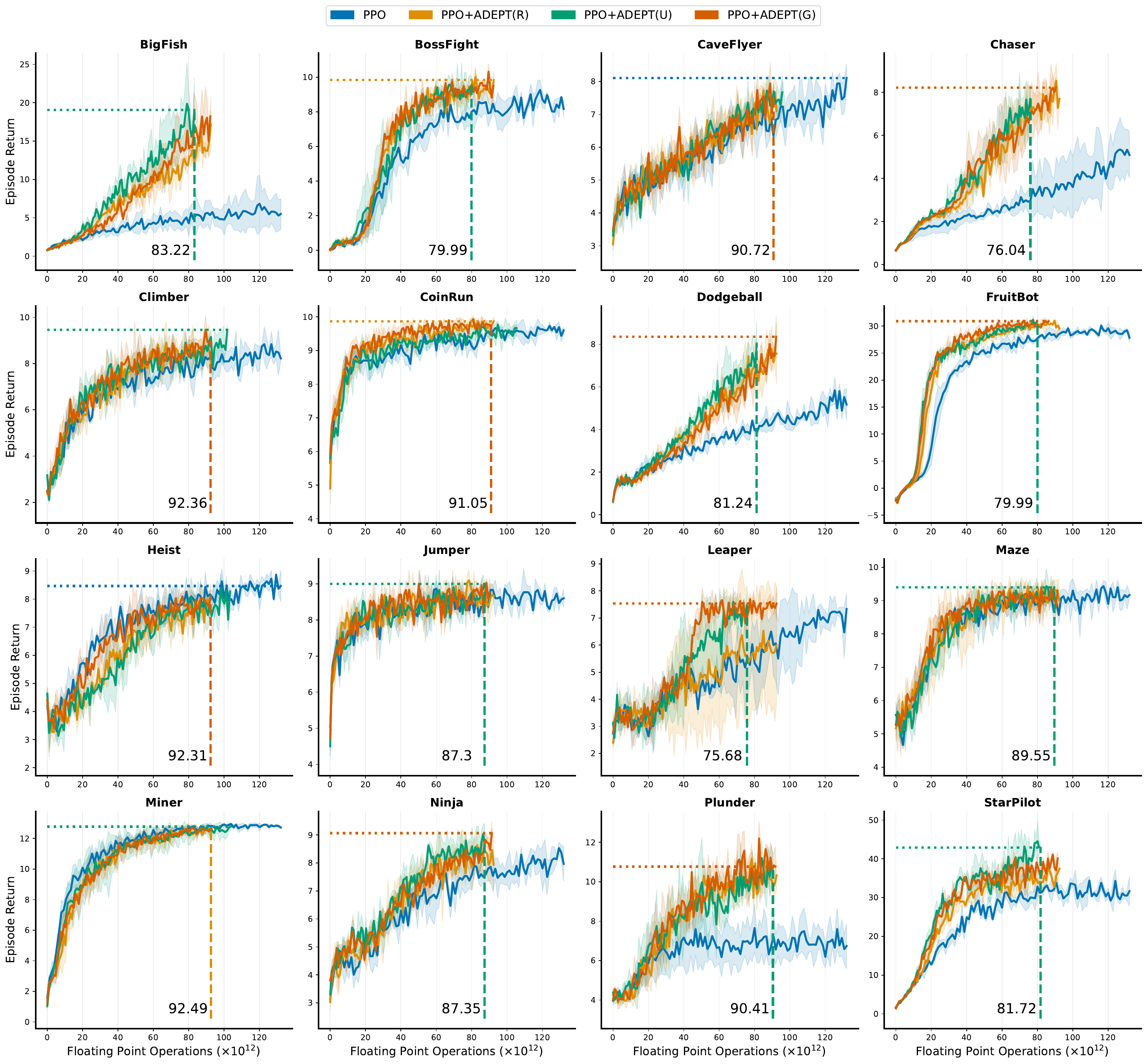}
\caption{Performance and overhead comparison of the vanilla PPO agent and its combinations with ADEPT on the Procgen benchmark. The solid line and shaded regions represent the mean and standard deviation, respectively, across five runs. Note that the dotted line and dashed line represent the highest score and the lowest overhead, respectively.}
\label{fig:pg_ppo_st16}
\vskip -0.2in
\end{figure*}

\clearpage\newpage

\begin{figure*}[h!]
\vskip 0.2in
\centering
\includegraphics[width=\linewidth]{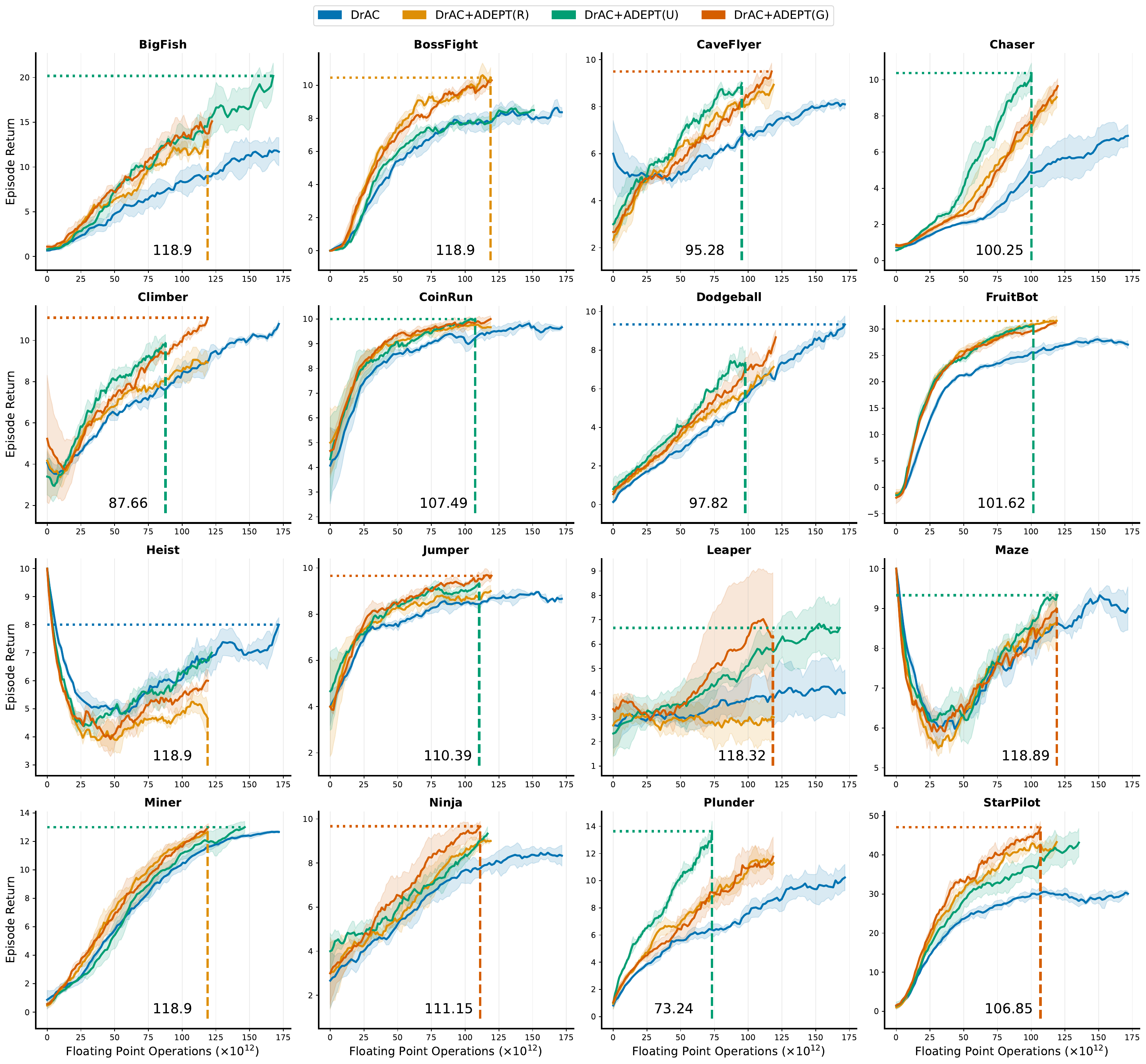}
\caption{Performance and overhead comparison of the vanilla DrAC agent and its combinations with ADEPT on the Procgen benchmark. The solid line and shaded regions represent the mean and standard deviation, respectively, across five runs. Note that the dotted line and dashed line represent the highest score and the lowest overhead, respectively.}
\label{fig:pg_drac_st16}
\vskip -0.2in
\end{figure*}

% \section{Data Efficiency Comparison}\label{appendix:efficiency}
% \begin{figure*}[h!]
% \vskip 0.2in
% \centering
% \includegraphics[width=\linewidth]{figures/pg_daac_st16.pdf}
% \caption{Performance and overhead comparison of the vanilla DAAC agent and its combinations with ADEPT on the Procgen benchmark. The solid line and shaded regions represent the mean and standard deviation, respectively, across five runs. Note that the dotted line and dashed line represent the highest score and the shortest time cost, respectively.}
% \label{fig:pg_daac_st16}
% \vskip -0.2in
% \end{figure*}

\clearpage\newpage

\section{Detailed Decision Processes}\label{appendix:decision}
\subsection{PPO+ADEPT(U)+Procgen}
\begin{figure*}[h!]
\vskip 0.2in
\centering
\includegraphics[width=\linewidth]{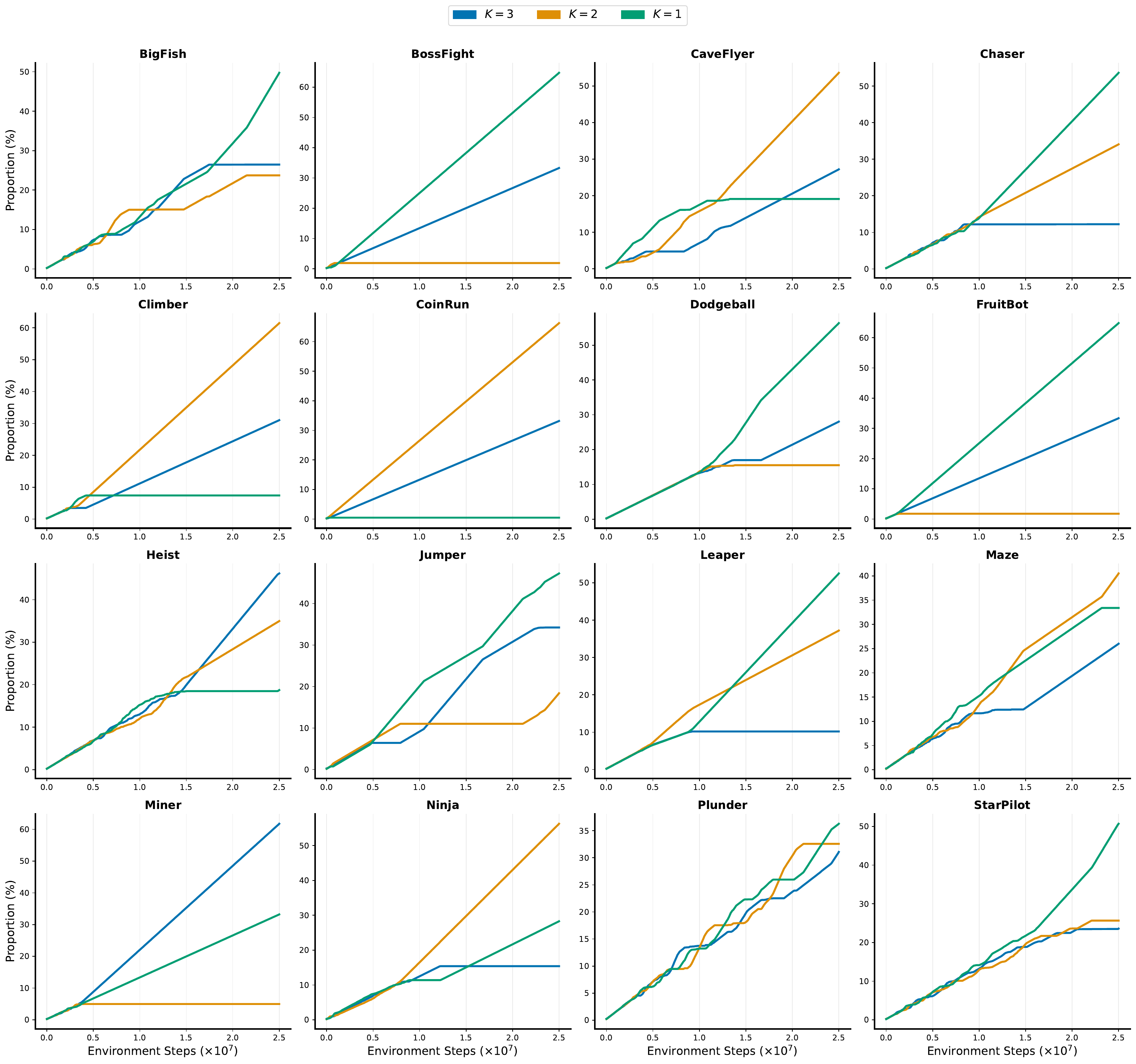}
\caption{Detailed decision processes of PPO+ADEPT(U) on the Procgen benchmark.}
\label{fig:pg_ppo_ucb_decision_16}
\vskip -0.2in
\end{figure*}
\clearpage\newpage

\subsection{PPO+ADEPT(G)+Procgen}
\begin{figure*}[h!]
\vskip 0.2in
\centering
\includegraphics[width=\linewidth]{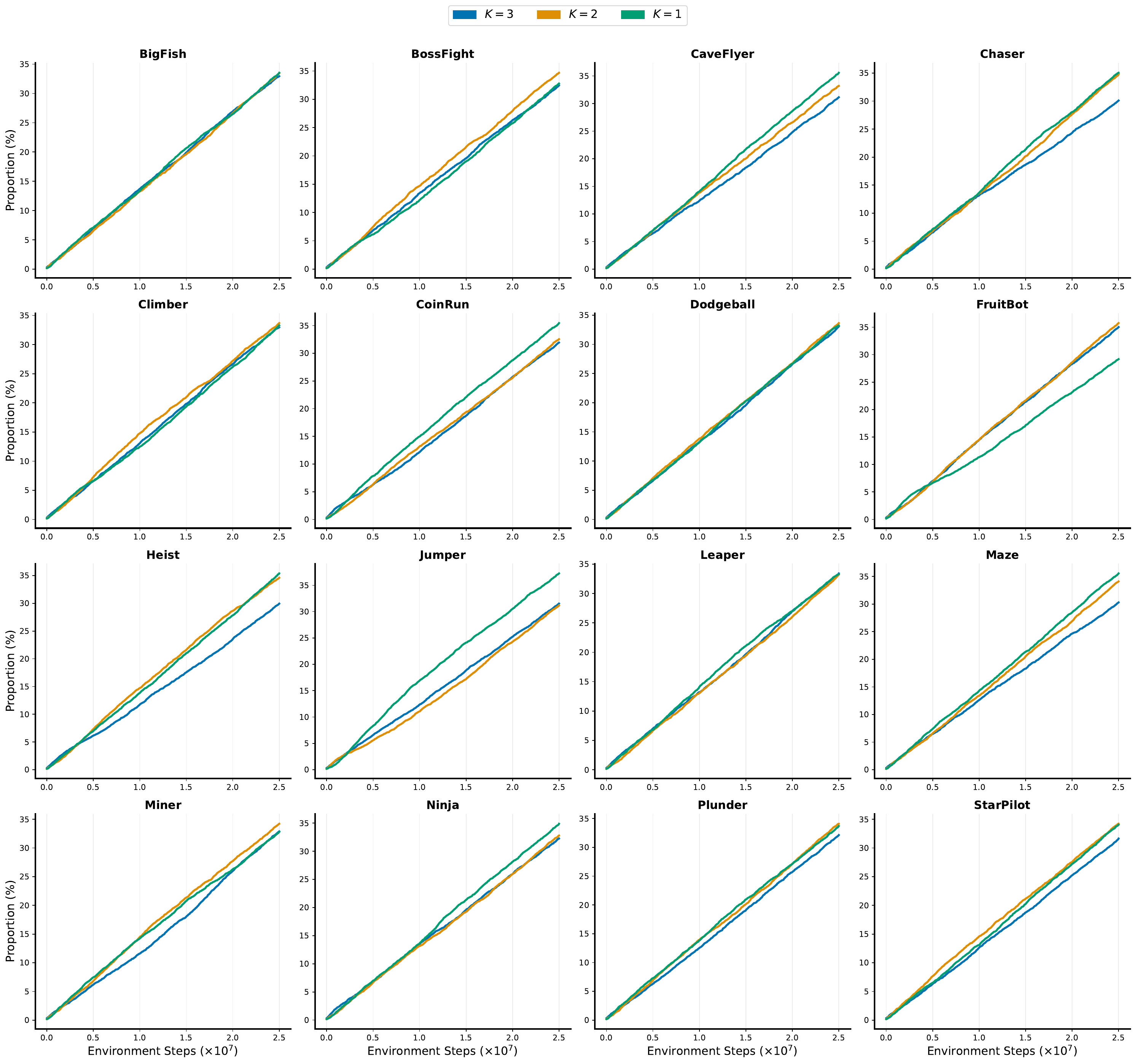}
\caption{Detailed decision processes of PPO+ADEPT(G) on the Procgen benchmark.}
\label{fig:pg_ppo_ts_decision_16}
\vskip -0.2in
\end{figure*}
\clearpage\newpage

\subsection{DrAC+ADEPT(U)+Procgen}
\begin{figure*}[h!]
\vskip 0.2in
\centering
\includegraphics[width=\linewidth]{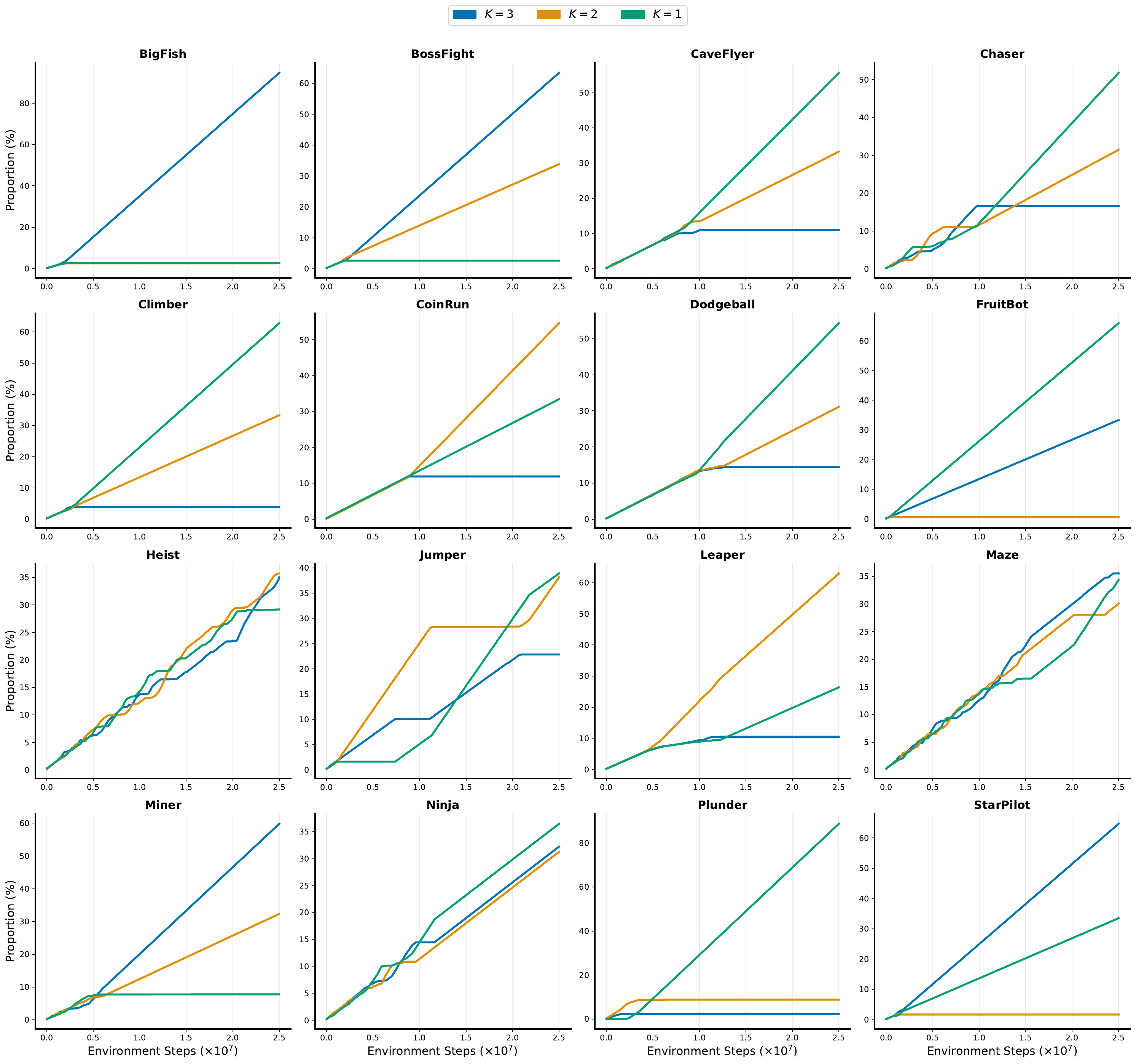}
\caption{Detailed decision processes of DrAC+ADEPT(U) on the Procgen benchmark.}
\label{fig:pg_drac_ucb_decision_16}
\vskip -0.2in
\end{figure*}
\clearpage\newpage

\subsection{DrAC+ADEPT(G)+Procgen}
\begin{figure*}[h!]
\vskip 0.2in
\centering
\includegraphics[width=\linewidth]{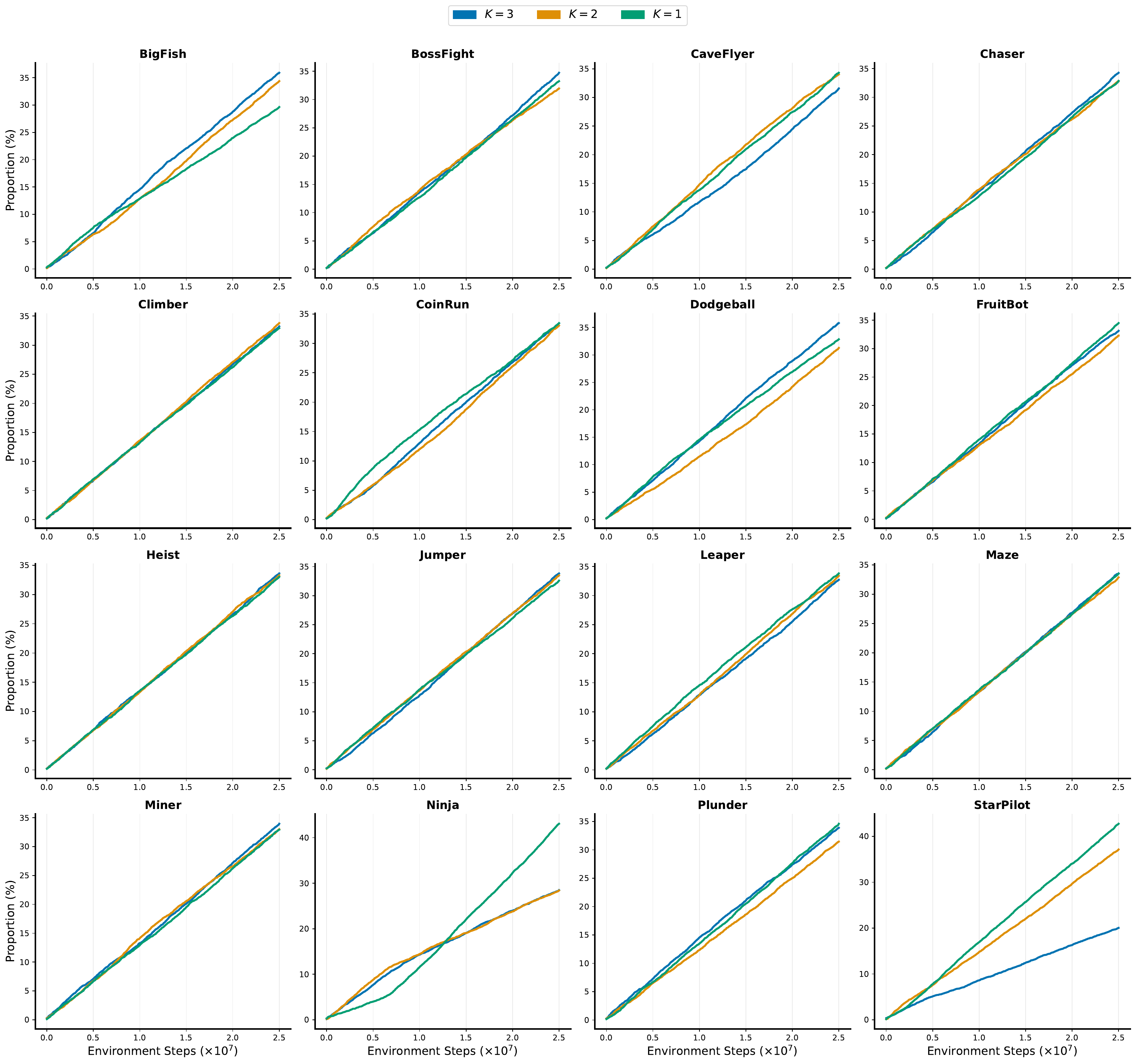}
\caption{Detailed decision processes of DrAC+ADEPT(G) on the Procgen benchmark.}
\label{fig:pg_drac_ts_decision_16}
\vskip -0.2in
\end{figure*}
\clearpage\newpage

% \subsection{DAAC+ADEPT(U)+Procgen}
% \begin{figure*}[h!]
% \vskip 0.2in
% \centering
% \includegraphics[width=\linewidth]{figures/pg_daac_ucb_decision_16.pdf}
% \caption{Detailed decision processes of DAAC+ADEPT(U) on the Procgen benchmark.}
% \label{fig:pg_daac_ucb_decision_16}
% \vskip -0.2in
% \end{figure*}
% \clearpage\newpage

% \subsection{DAAC+ADEPT(G)+Procgen}
% \begin{figure*}[h!]
% \vskip 0.2in
% \centering
% \includegraphics[width=\linewidth]{figures/pg_daac_ts_decision_16.pdf}
% \caption{Detailed decision processes of DAAC+ADEPT(G) on the Procgen benchmark.}
% \label{fig:pg_daac_ts_decision_16}
% \vskip -0.2in
% \end{figure*}
% \clearpage\newpage

\section{Ablation Studies}\label{appendix:ablation}
\subsection{Hyperparameter Search}\label{appendix:ablation hp}
\subsubsection{PPO+ADEPT(U)+Procgen}
\begin{figure}[h!]
\vskip 0.2in
\begin{center}
\centerline{\includegraphics[width=0.6\linewidth]{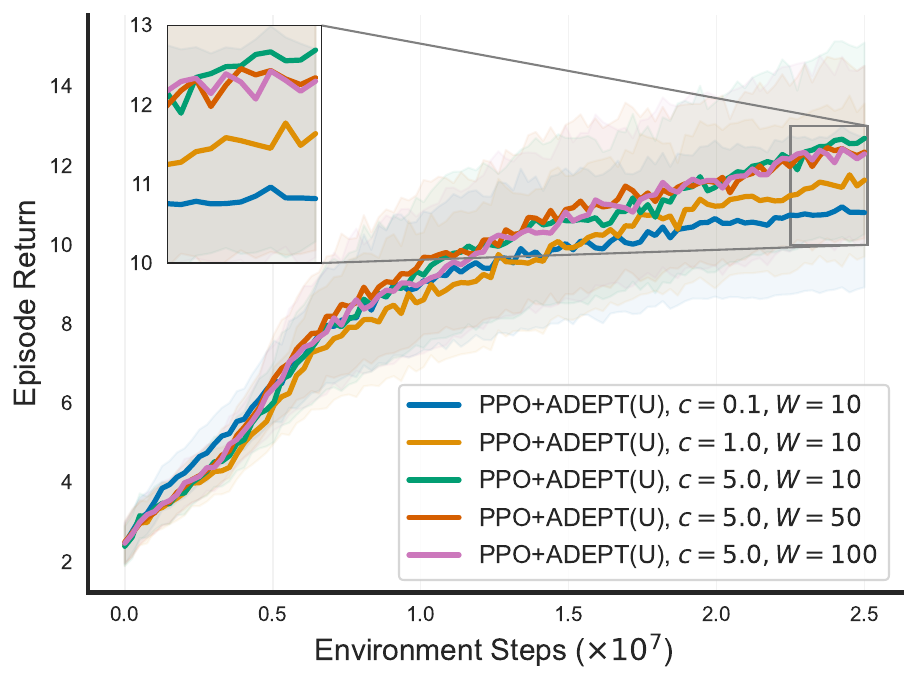}}
\caption{Aggregated training performance comparison of PPO+ADEPT(U) with different exploration coefficients and sizes of the sliding window. The mean and standard deviation are computed across all the environments.}
\label{fig:pg_ppo_ucb_ablations}
\end{center}
\vskip -0.2in
\end{figure}

\subsubsection{PPO+ADEPT(G)+Procgen}

\begin{figure}[h!]
\vskip 0.2in
\begin{center}
\centerline{\includegraphics[width=0.6\linewidth]{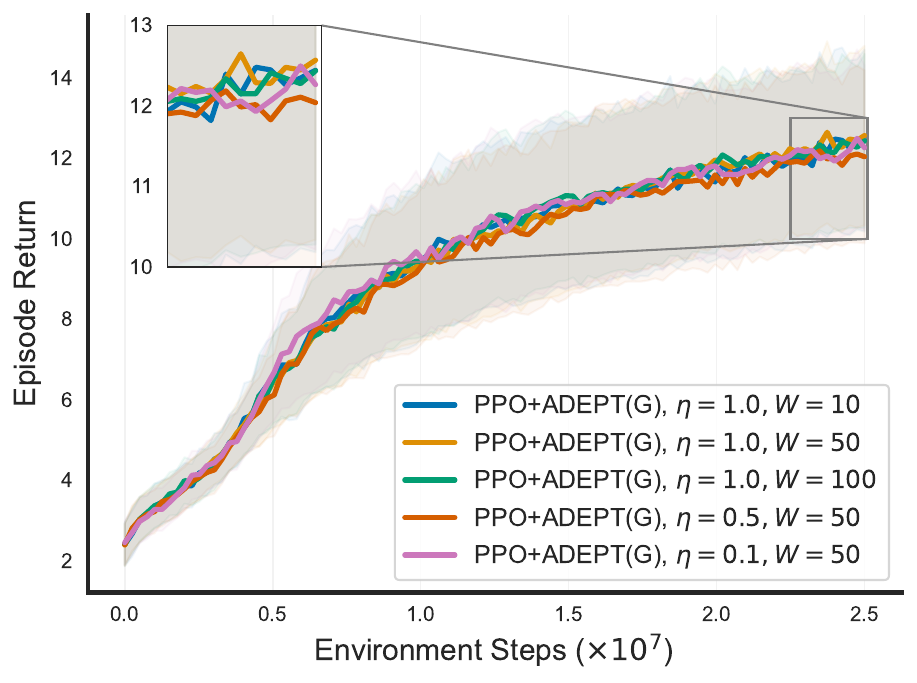}}
\caption{Aggregated training performance comparison of PPO+ADEPT(G) with different learning rates and sizes of the sliding window. The mean and standard deviation are computed across all the environments.}
\label{fig:pg_ppo_ts_ablations}
\end{center}
\vskip -0.2in
\end{figure}

\clearpage\newpage

\subsubsection{DrAC+ADEPT(U)+Procgen}

\begin{figure}[h!]
\vskip 0.2in
\begin{center}
\centerline{\includegraphics[width=0.6\linewidth]{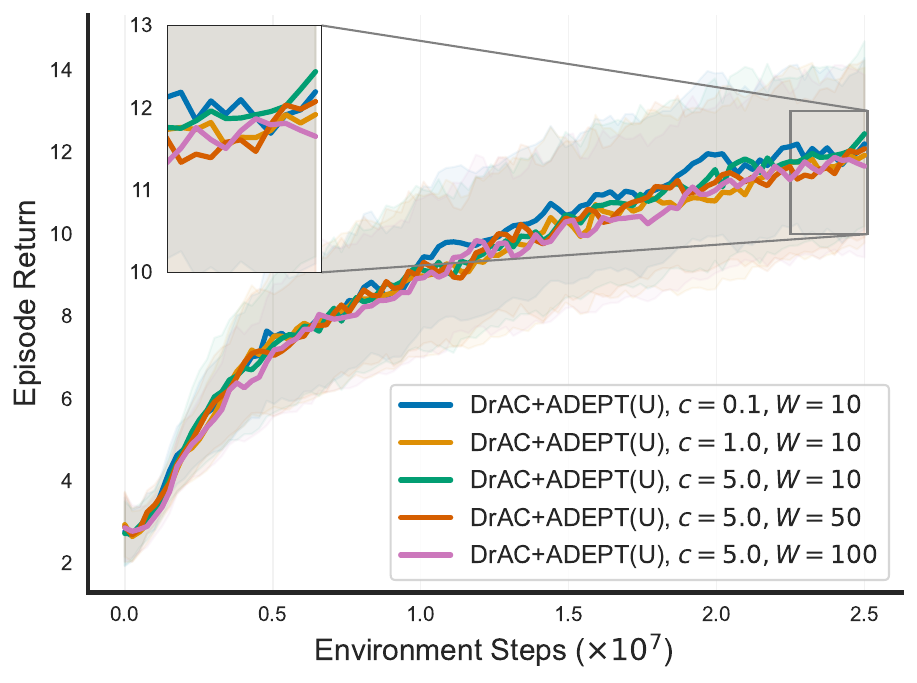}}
\caption{Aggregated training performance comparison of DrAC+ADEPT(U) with different exploration coefficients and sizes of the sliding window. The mean and standard deviation are computed across all the environments.}
\label{fig:pg_drac_ucb_ablations}
\end{center}
\vskip -0.2in
\end{figure}

\subsubsection{DrAC+ADEPT(G)+Procgen}

\begin{figure}[h!]
\vskip 0.2in
\begin{center}
\centerline{\includegraphics[width=0.6\linewidth]{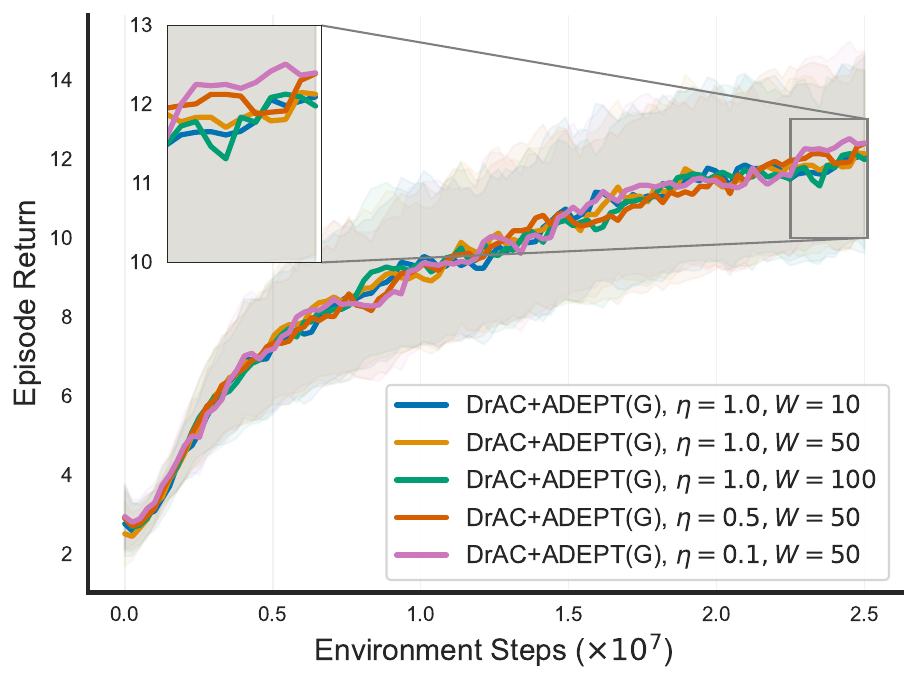}}
\caption{Aggregated training performance comparison of DrAC+ADEPT(G) with different learning rates and sizes of the sliding window. The mean and standard deviation are computed across all the environments.}
\label{fig:pg_drac_ts_ablations}
\end{center}
\vskip -0.2in
\end{figure}

\clearpage\newpage

\subsection{Different NUE Sets}\label{appendix:ablation nue}
\subsubsection{PPO+ADEPT(R)+Procgen}
\begin{figure}[h!]
\vskip 0.2in
\begin{center}
\centerline{\includegraphics[width=\linewidth]{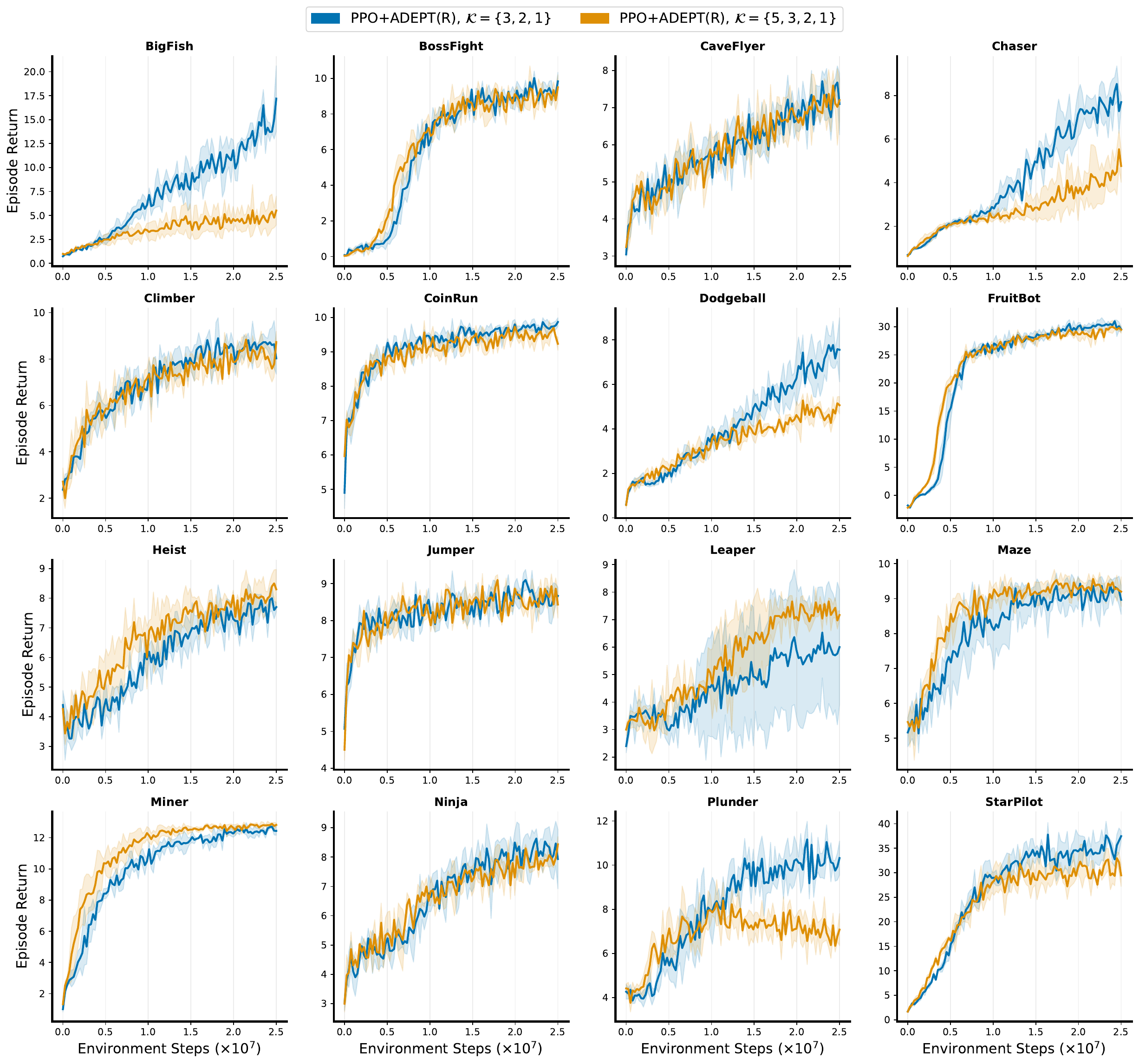}}
\caption{Aggregated training performance comparison of PPO+ADEPT(R) with different sets of NUE values. The mean and standard deviation are computed across all the environments.}
\label{fig:pg_ppo_rr_nue}
\end{center}
\vskip -0.2in
\end{figure}

\clearpage\newpage

\subsubsection{PPO+ADEPT(U)+Procgen}
\begin{figure}[h!]
\vskip 0.2in
\begin{center}
\centerline{\includegraphics[width=\linewidth]{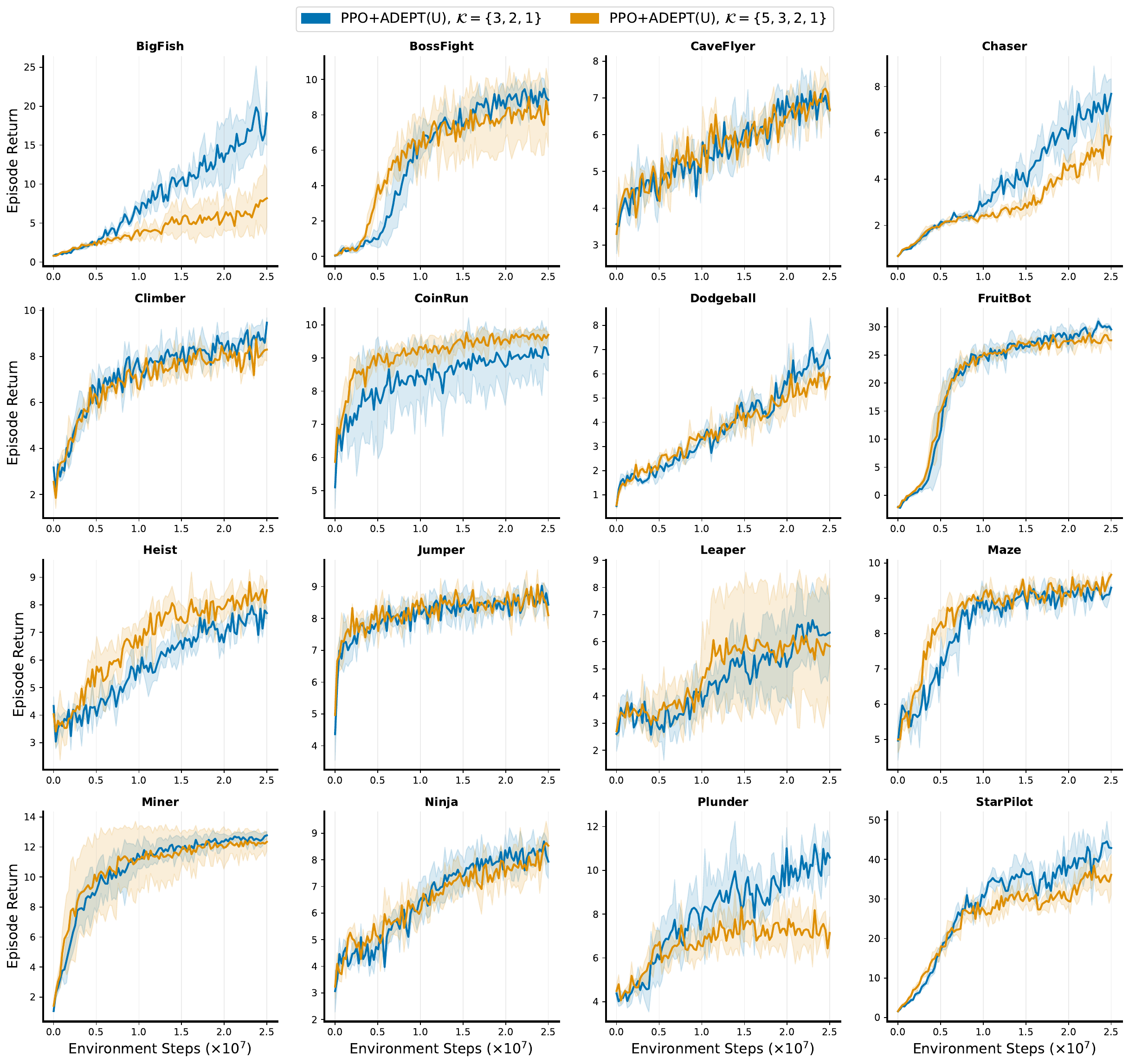}}
\caption{Aggregated training performance comparison of PPO+ADEPT(U) with different sets of NUE values. Here, the exploration coefficient $c$ is $5.0$, and the length $W$ of the sliding window is $10$. The mean and standard deviation are computed across all the environments.}
\label{fig:pg_ppo_ucb_nue}
\end{center}
\vskip -0.2in
\end{figure}

\clearpage\newpage

\subsubsection{PPO+ADEPT(G)+Procgen}
\begin{figure}[h!]
\vskip 0.2in
\begin{center}
\centerline{\includegraphics[width=\linewidth]{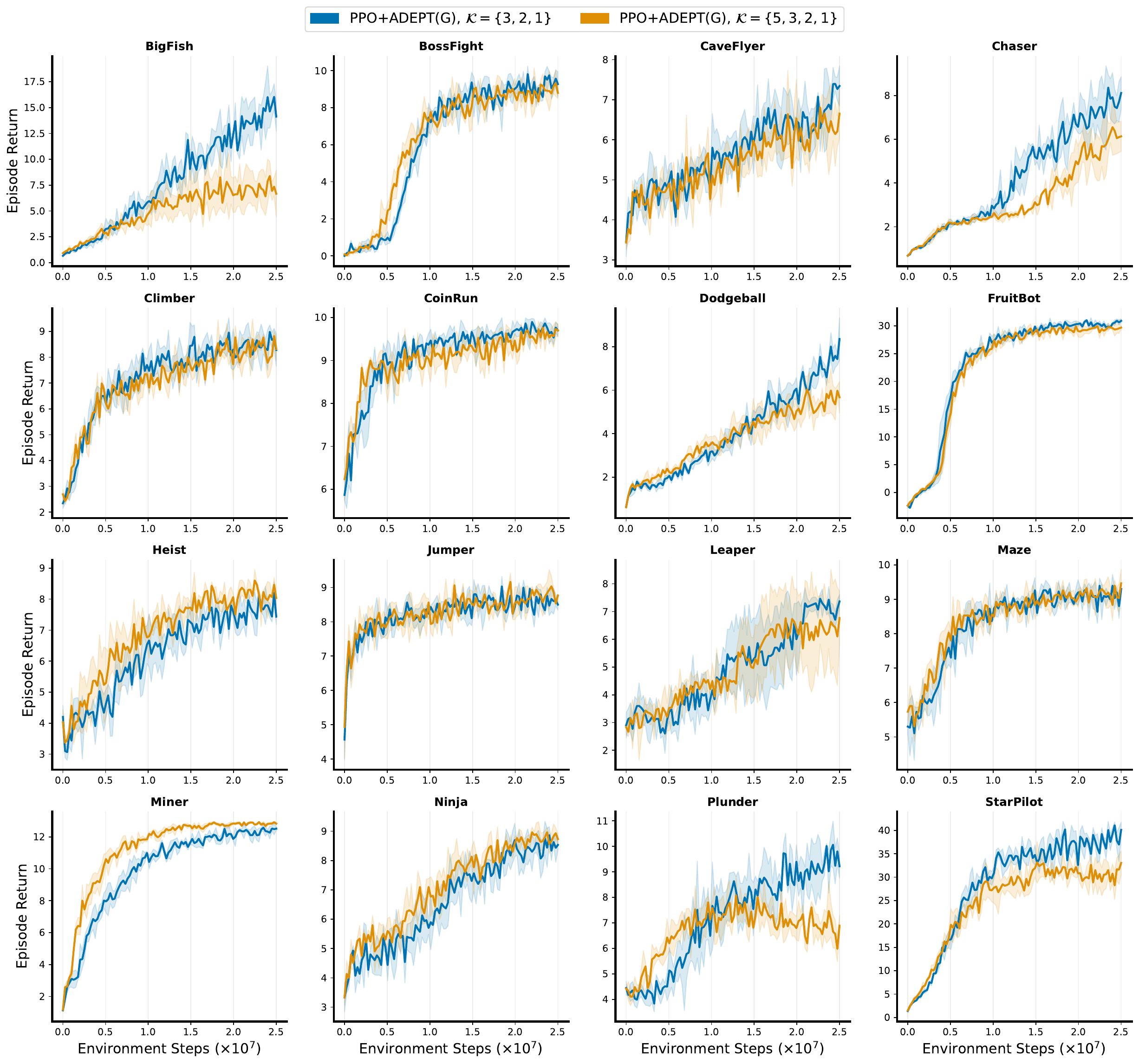}}
\caption{Aggregated training performance comparison of PPO+ADEPT(G) with different sets of NUE values. Here, the learning rate $\eta$ is $1.0$, and the length $W$ of the sliding window is $10$. The mean and standard deviation are computed across all the environments.}
\label{fig:pg_ppo_ts_nue}
\end{center}
\vskip -0.2in
\end{figure}

\clearpage\newpage

\subsubsection{DrAC+ADEPT(R)+Procgen}
\begin{figure}[h!]
\vskip 0.2in
\begin{center}
\centerline{\includegraphics[width=\linewidth]{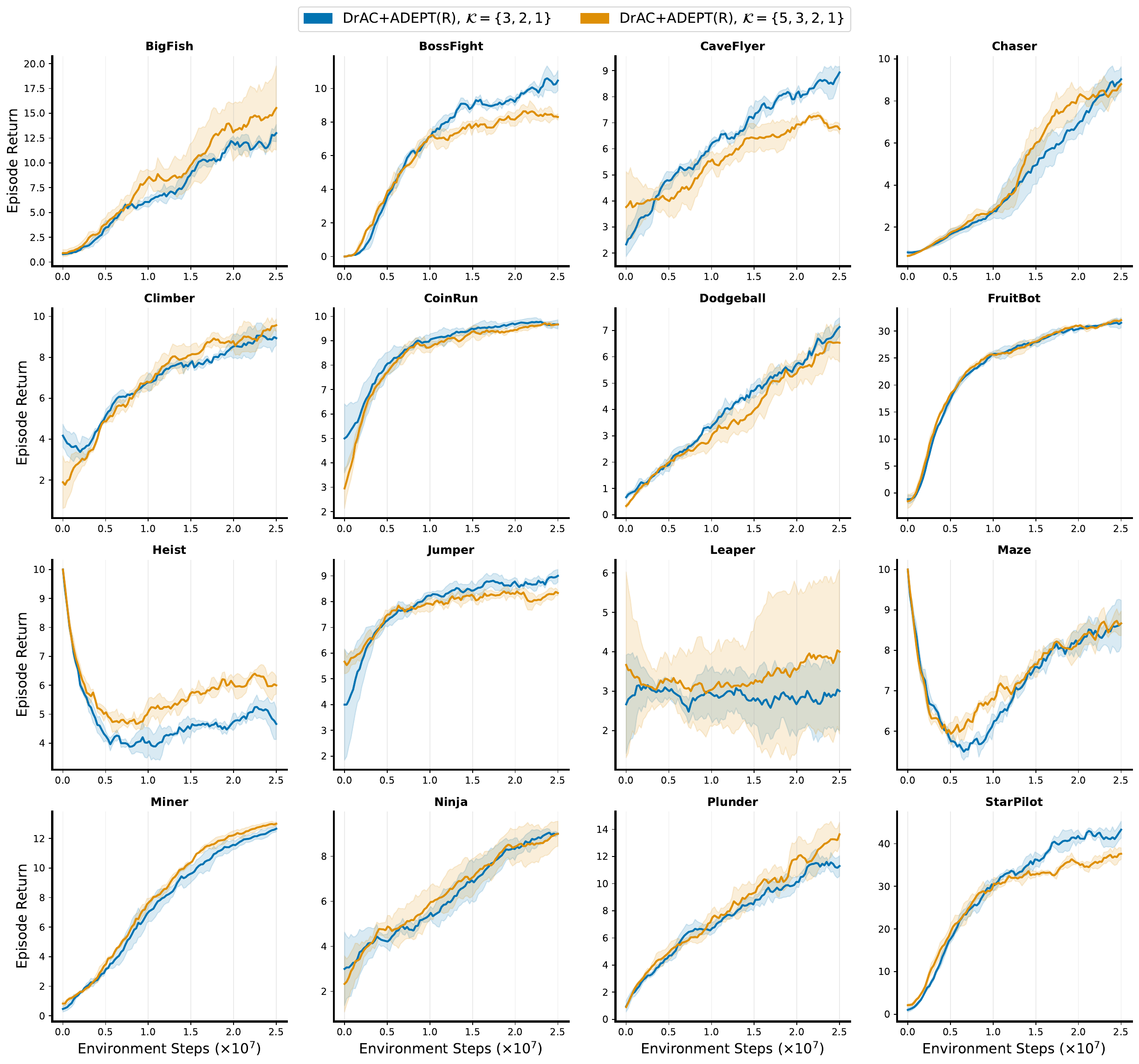}}
\caption{Aggregated training performance comparison of DrAC+ADEPT(R) with different sets of NUE values. The mean and standard deviation are computed across all the environments.}
\label{fig:pg_drac_rr_nue}
\end{center}
\vskip -0.2in
\end{figure}

\clearpage\newpage

\subsubsection{DrAC+ADEPT(U)+Procgen}
\begin{figure}[h!]
\vskip 0.2in
\begin{center}
\centerline{\includegraphics[width=\linewidth]{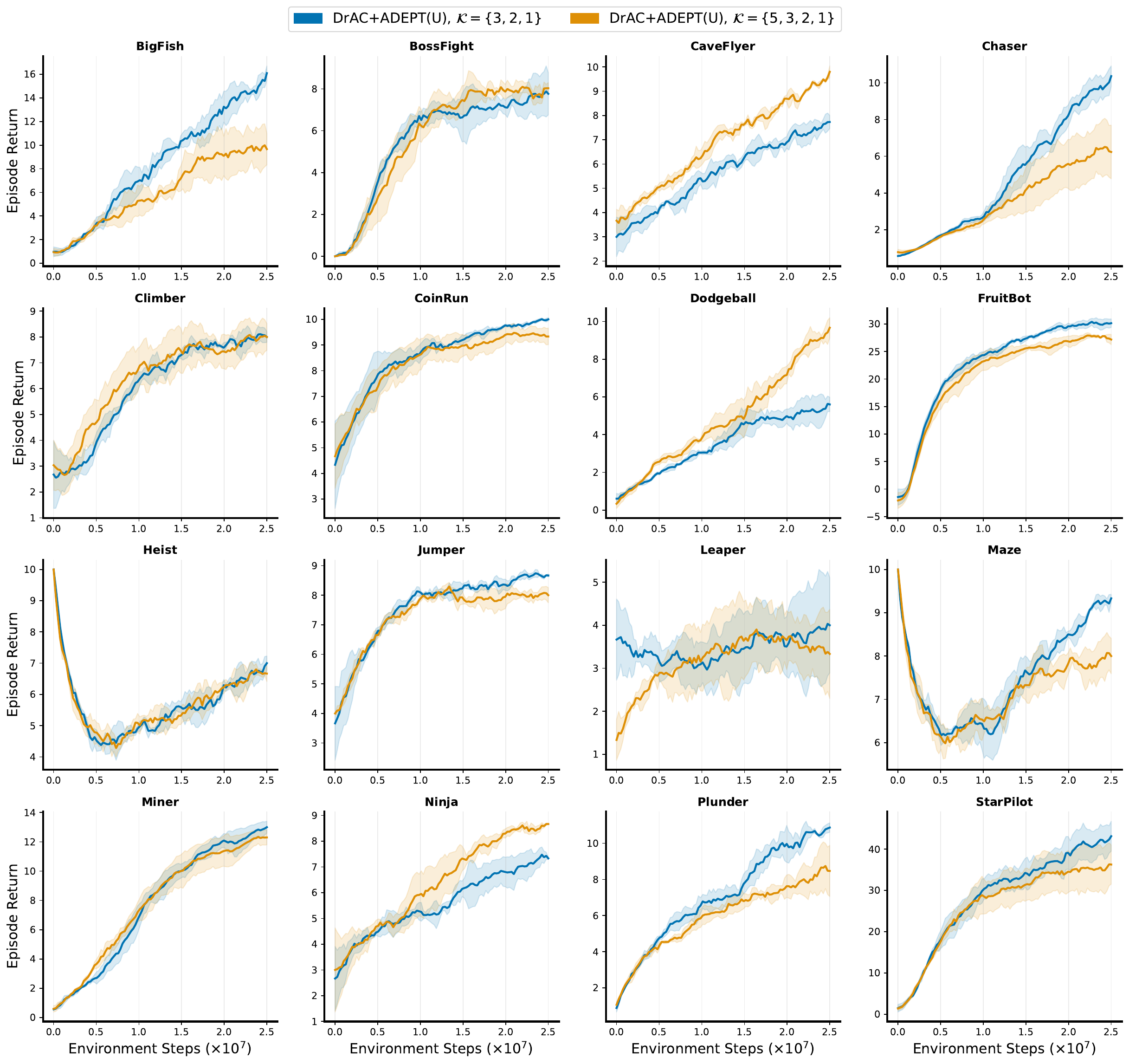}}
\caption{Aggregated training performance comparison of DrAC+ADEPT(U) with different sets of NUE values. Here, the exploration coefficient $c$ is $5.0$, and the length $W$ of the sliding window is $10$. The mean and standard deviation are computed across all the environments.}
\label{fig:pg_drac_ucb_nue}
\end{center}
\vskip -0.2in
\end{figure}

\clearpage\newpage

\subsubsection{DrAC+ADEPT(G)+Procgen}
\begin{figure}[h!]
\vskip 0.2in
\begin{center}
\centerline{\includegraphics[width=\linewidth]{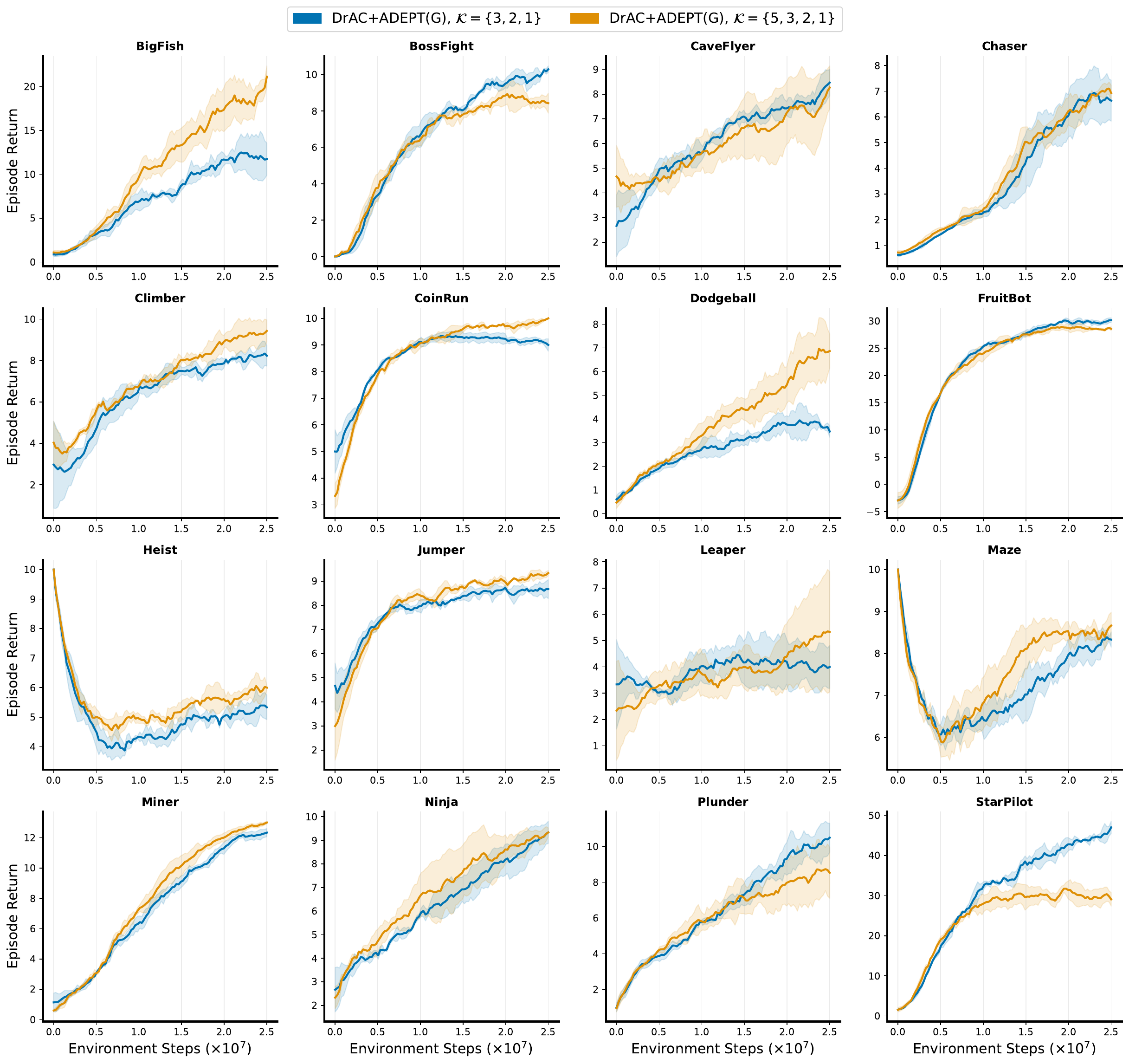}}
\caption{Aggregated training performance comparison of DrAC+ADEPT(G) with different sets of NUE values. Here, the learning rate $\eta$ is $1.0$, and the length $W$ of the sliding window is $10$. The mean and standard deviation are computed across all the environments.}
\label{fig:pg_drac_ts_nue}
\end{center}
\vskip -0.2in
\end{figure}

% \begin{figure}[h!]
% \vskip 0.2in
% \begin{center}
% \centerline{\includegraphics[width=0.6\linewidth]{figures/pg_daac_ucb_ablations.pdf}}
% \caption{Aggregated performance comparison of ADEPT(U)+DAAC with different sizes of the sliding window. The mean and standard deviation are computed across all the environments.}
% \label{fig:pg_daac_ucb_ablations}
% \end{center}
% \vskip -0.2in
% \end{figure}

\clearpage\newpage
\section{Calculation of Computational Overhead}\label{appendix:overhead}
To compare the computational efficiency of the baseline algorithms and ADEPT, we utilize the floating point operations (FLOPS) as the KPI. Moreover, we only count the computational overhead of the network-involved operations, such as the data sampling and model update. In the Procgen experiments, all the methods use an identical architecture for the policy and the value network, which can be found in \cite{cobbe2020leveraging}. We leverage an open-source tool entitled \textbf{PyTorch-OpCounter} \cite{thop} to calculate its FLOPS, and the result is denoted as $O_{\rm bs1}$ for batch size=1.

For the sampling phase, the computational overhead is 
\begin{equation}
    O_{\rm sampling}=(N_{\rm episode\;length}+1)*N_{\rm environments}*O_{\rm bs1}
\end{equation}
% \begin{equation}
%     {\rm sampling\_overhead}=({\rm num\_steps}+1) * {\rm num\_envs} * {\rm flops\_bs\_1},
% \end{equation}
where $+1$ is for predicting the returns of the next observations at the end of the episode, as shown in the PPO implementation of CleanRL \cite{huang2022cleanrl}.

For the model update phase, the computational overhead is 
\begin{equation}
    O_{\rm update}=O_{\rm forward}+O_{\rm backward},
\end{equation}
where 
\begin{equation}
    O_{\rm forward}=O_{\rm bs1} * B * N_{\rm batches} * N_{\rm update\;epochs}
\end{equation}
and
\begin{equation}
    O_{\rm backward}=O_{\rm forward}*2.
\end{equation}
Here, $B$ is the batch size, and the overhead ratio of a forward pass to a backward pass is \textbf{1:2} as suggested by \cite{amodei2019ai}.

Finally, the total computational overhead is 
\begin{equation}
    O_{\rm total}=(O_{\rm sampling}+O_{\rm update})*N_{\rm episodes}
\end{equation}

In the Procgen experiments, we have
\begin{equation}
\begin{aligned}
    O_{\rm bs1}&=528384\mathrm{FLOPS}\\
    N_{\rm environments}&=64\\
    N_{\rm episode\;length}&=256\\
    N_{\rm episodes}&=1525\\
    B&=2048\\
    N_{\rm batches}&=32
\end{aligned}    
\end{equation}

\end{document}